\documentclass[lettersize,journal]{IEEEtran}
\usepackage{amsmath,amsfonts}
\usepackage{algorithmic}
\usepackage{array}
\usepackage[caption=false,font=normalsize,labelfont=sf,textfont=sf]{subfig}
\usepackage{textcomp}
\usepackage{stfloats}
\usepackage{url}
\usepackage{verbatim}
\usepackage{graphicx}
\def\BibTeX{{\rm B\kern-.05em{\sc i\kern-.025em b}\kern-.08em
    T\kern-.1667em\lower.7ex\hbox{E}\kern-.125emX}}
\usepackage{balance}

\usepackage{multirow}
\usepackage{placeins} 
\usepackage[utf8]{inputenc}
\DeclareUnicodeCharacter{200B}{ }
\usepackage{soul,xcolor, colortbl} 

\usepackage{algorithmic}
\usepackage{algorithm}
\usepackage{amsmath}
\usepackage{mathtools}
\usepackage{amsfonts}
\usepackage{amsthm}
\usepackage{amssymb}
\usepackage{listings}
\usepackage{color}
\usepackage{graphicx}
\definecolor{dkgreen}{rgb}{0,0.6,0}
\definecolor{gray}{rgb}{0.5,0.5,0.5}
\definecolor{mauve}{rgb}{0.58,0,0.82}
\usepackage[left=2.0cm,right=2.0cm,top=0.7cm,bottom=0.5cm,includehead,includefoot]{geometry}

\begin{document}
\title{ALSS-YOLO: An Adaptive Lightweight Channel Split and Shuffling Network for TIR Wildlife Detection in UAV Imagery}

\author{Ang He, Xiaobo Li, Ximei Wu, Chengyue Su, Jing Chen, Sheng Xu*, Xiaobin Guo*

\thanks{This work was supported by the National Natural Science Foundation of China (No. 11904056). Guangzhou Basic and Applied Basic Research Project (No. 202102020053).}
\thanks{Guangdong Provincial Key Laboratory of Sensing Physics and System Integration Applications, Guangdong University of Technology, Guangzhou Higher Education Mega Centre, Guangzhou, 510006, PR China; 2112215151@mail2.gdut.edu.cn (A.-H.);  }
\thanks{*Correspondence: xusheng@gdut.edu.cn(S.-X.) guoxb@gdut.edu.cn (X.-b.G.)}}

\markboth{Journal of \LaTeX\ Class Files,~Vol.~18, No.~9, September~2020}%
{How to Use the IEEEtran \LaTeX \ Templates}
\maketitle

\begin{abstract}
Unmanned aerial vehicles (UAVs) equipped with thermal infrared (TIR) cameras play a crucial role in combating nocturnal wildlife poaching. However, TIR images often face challenges such as jitter, and wildlife overlap, necessitating UAVs to possess the capability to identify blurred and overlapping small targets. Current traditional lightweight networks deployed on UAVs struggle to extract features from blurry small targets. To address this issue, we developed ALSS-YOLO, an efficient and lightweight detector optimized for TIR aerial images. Firstly, we propose a novel Adaptive Lightweight Channel Split and Shuffling (ALSS) module. This module employs an adaptive channel split strategy to optimize feature extraction and integrates a channel shuffling mechanism to enhance information exchange between channels.  This improves the extraction of blurry features, crucial for handling jitter-induced blur and overlapping targets. Secondly, we developed a Lightweight Coordinate Attention (LCA) module that employs adaptive pooling and grouped convolution to integrate feature information across dimensions. This module ensures lightweight operation while maintaining high detection precision and robustness against jitter and target overlap. Additionally, we developed a single-channel focus module to aggregate the width and height information of each channel into four-dimensional channel fusion, which improves the feature representation efficiency of infrared images. Finally, we modify the localization loss function to emphasize the loss value associated with small objects to improve localization accuracy. Extensive experiments on the BIRDSAI and ISOD TIR UAV wildlife datasets show that ALSS-YOLO achieves state-of-the-art performance, Our code is openly available at https://github.com/helloworlder8/computer\_vision.

\end{abstract}

\begin{IEEEkeywords}
unmanned aerial vehicles (UAVs), thermal infrared (TIR), wildlife detection, small targets, lightweight detector.
\end{IEEEkeywords}

\section{Introduction}
\IEEEPARstart{E}{fficient} wildlife management relies on frequent population monitoring. Traditional approaches, such as patrolling protected areas, particularly during nighttime, encounter significant obstacles. These methods demand extensive time, and substantial financial resources, and subject forest rangers to risks arising from reduced visibility, rugged terrain, and heightened poacher activities. With technological advancements, manned aircraft have gradually replaced traditional methods\cite{int_01_01}, Nevertheless, limitations remain due to high operational costs and safe flight restrictions. Recently, unmanned aerial vehicles (UAVs) have emerged as a viable option for wildlife and habitat management. UAVs can access hazardous areas inaccessible or difficult to approach by humans, with lower operational costs and greater maneuverability\cite{int_01_02}\cite{int_01_03}.

\begin{figure}[hptb]
\includegraphics[width=8.5 cm]{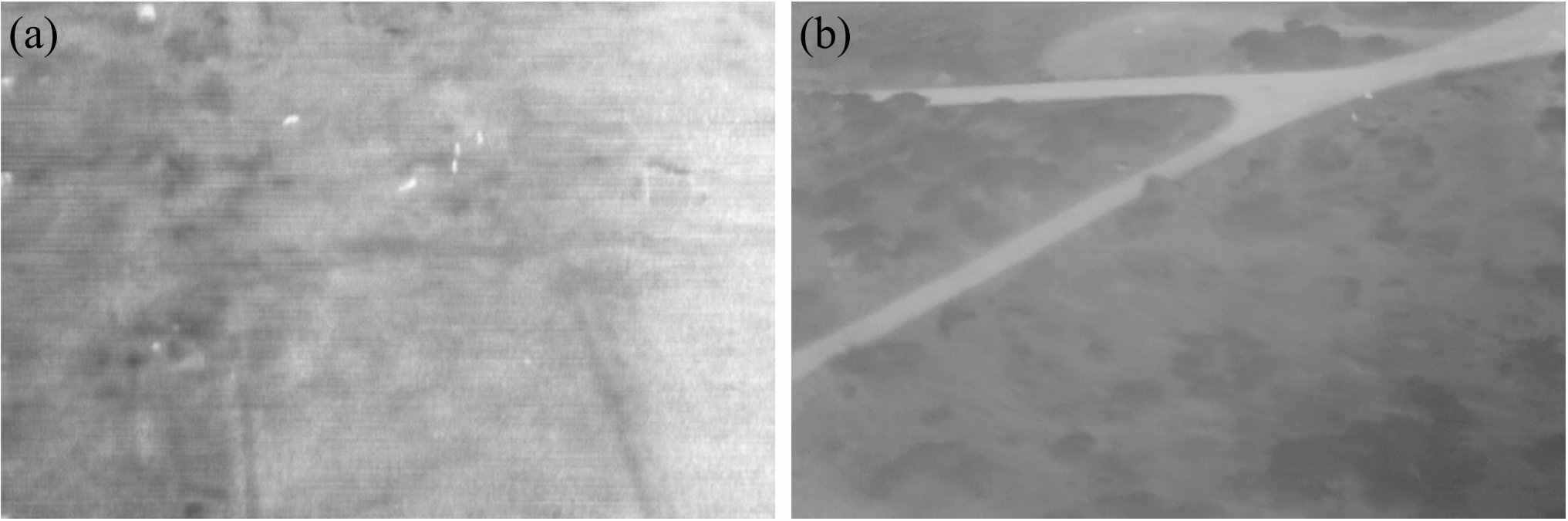}
\caption{Examples of blurred FIR wildlife data photos included in the BIRDSAI dataset\cite{BIRDSAI} by noise or weather conditions. (\textbf{a}) Blurred by noise. (\textbf{b}) Blurred by weather conditions.
\label{BIRDSAI-examples}}
\end{figure}

Visible light imagery stands as the most prevalent data format collected by UAVs\cite{int_02_01}\cite{int_02_02}. Conversely, thermal infrared (TIR) imaging sensors offer valuable insights compared to color imaging sensors. In recent research, a customized ERGW-net\cite{int_02_03} was designed and fine-tuned to detect small roadside objects using a TIR camera for 24-hour monitoring. Recognizing the potential of TIR imaging for detecting small objects in challenging environments, we designed ALSS-YOLO with the need to enhance the detection capabilities of small objects in mind. 
TIR imagery has traditionally been utilized for medical, security, military, and autonomous driving\cite{int_02_04} purposes. In recent years, due to their low cost and high image quality, TIR cameras have become the preferred sensors for nighttime surveillance in natural observation. In scenarios with limited visibility, the height, payload capacity, and stealth requirements of UAVs preclude the use of visible light sensors. Integration of UAVs with TIR cameras has been demonstrated as highly conducive to animal monitoring in intricate wildlife environments.\cite{int_02_05}\cite{int_02_06}\cite{CE-RetinaNet}.
However, due to its unique characteristics, object detection in TIR images remains challenging. TIR images typically exhibit low resolution and significant noise.\cite{int_03_01}. Compared to color images, TIR images have only one channel, resulting in the loss of visual details, and only the contours are preserved. In Ref.\cite{BIRDSAI}, the collected dataset of infrared wildlife images includes examples of photos blurred due to noise or weather conditions, as shown in Fig.~\ref{BIRDSAI-examples}. Additionally, changes in the drone's flight altitude and camera field of view can cause large changes in the apparent scale of the same target. besides, from the UAV's perspective, targets often appear as dense small objects, which increases the rate of false alarms during detection. For these reasons, even human experts often struggle to accurately identify wildlife, leading to recognition errors.  
To address these issues, we designed the Adaptive Lightweight Channel Split and Shuffling (ALSS) and Lightweight Coordinate Attention (LCA) modules and developed the FineSIOU loss function. These components were specifically engineered to enhance detection precision and speed, effectively overcoming the limitations identified in previous studies.

\unskip

Venkatachalam et al.\cite{venkatachalam} designed a specialized infrared wildlife detector using an optimized Region-based Convolutional Neural Network (R-CNN) model. By employing transfer learning and fine-tuning on a smaller annotated infrared image dataset, the accuracy of animal region detection and segmentation was significantly enhanced. However, the model's complexity renders it unsuitable for real-time detection on UAV platforms.
Existing lightweight infrared drone detection algorithms, such as PHSI-RTDETR\cite{PHSI_RTDETR}, have demonstrated notable efficacy in detecting small infrared targets. An exemplary model in this domain, SLBAF-Net\cite{SLBAF_Net}, is an ultra-lightweight bimodal network designed for drone detection under challenging lighting and meteorological conditions. SLBAF-Net adaptively fuses visible light and infrared images, incorporating a bimodal adaptive fusion module (BAFM) inspired by the YOLO network architecture to enhance detection robustness. However, these algorithms generally do not address severe occlusion and overlapping scenes adequately. Moreover, there is a lack of discussion regarding their performance in infrared wildlife detection scenarios.

This paper aims to address the limitations of current lightweight infrared detection algorithms in the presence of severe occlusion and overlapping targets, specifically in wildlife detection. Consequently, there is a need for a reliable and efficient object detector for TIR images on UAV platforms. To tackle these challenges, we introduce ALSS-YOLO, an effective wildlife detection solution based on deep convolutional neural networks (CNNs). The main contributions of this study are as follows:

\begin{enumerate}

\item	This study introduces innovative modules to enhance the network's feature extraction and representation capabilities while optimizing computational efficiency. We propose an ALSS module and a LCA module. 
The ALSS module uses an adaptive channel split strategy to optimize feature extraction and employs channel shuffling to enhance feature representation by increasing inter-channel information flow, thereby improving the detection accuracy of blurry targets. The LCA module uses adaptive pooling and grouped convolutions to enhance feature extraction along height and width dimensions, improving spatial information integration. This is especially useful for low-resolution TIR images, enhancing feature extraction and detection precision.
Additionally, we developed a single-channel focus module that aggregates width and height information into a four-dimensional channel fusion, converting spatial information into channel dimensions for better feature extraction.
\item	FineSIOU is designed as a localization loss function, separately handling angle costs in total loss computation. Shape loss adjustments emphasize the size and shape of ground truth bounding boxes, thereby enhancing the bounding box regression speed and TIR small object detection capabilities.
\item  Experiments conducted on both the BIRDSAI and ISOD datasets demonstrate the superiority of our proposed ALSS-YOLO algorithm. Specifically, on the BIRDSAI dataset, ALSS-YOLO (1.452 million parameters) achieves a 1.7\% higher mAP0.50 compared to YOLOv8-n' (1.795 million parameters). Additionally, we introduce the ALSS-YOLO-s (2.226 million parameters, 0.895 mAP0.50) and ALSS-YOLO-m (2.924 million parameters, 0.903 mAP0.50) models, both of which exhibit significant improvements in mAP scores and parameter efficiency when compared to other lightweight object detectors.
\end{enumerate}

The rest of this paper is organized as follows: Section II presents previous work on target detection, including generic target detection, TIR UAV target detection, small target detection, and lightweight network design. Section III describes the proposed network in detail. Section IV presents the experimental results and further discussion. Section V presents conclusions and future work.

\section{Related Work}

\subsection{Generic Object Detection}
Early object detection models were built by integrating a series of manually designed feature extractors. (such as Viola-Jones\cite{Viola-Jones} and HOG\cite{HOG}, etc.). These models are characterized due to their slow processing speeds, low accuracy, and limited generalizability. The breakthrough of convolutional neural networks in the field of target detection in the last decade has gradually replaced the traditional methods. The object detection algorithms can be categorized into two types: two-stage object detection algorithms, which involve generating candidate boxes and classifying objects within them, and single-stage object detection algorithms, which do not generate candidate boxes.
The former, such as R-CNN\cite{RCNN}, utilizes the Selective Search algorithm to generate approximately 2000 region proposals. Fast R-CNN\cite{Fast} and Faster R-CNN\cite{Faster} further introduced RoIPooling and RPN based on R-CNN, respectively. Subsequent advancements in target detection algorithms have focused on enhancing accuracy, with notable examples such as Mask R-CNN \cite{Mask}, HyperNet\cite{HyperNet}, and PVANet\cite{PVANet}. While these algorithms exhibit robustness and low error rates, they exhibit heavy computational loads, making them unsuitable for real-time applications. In contrast, single-stage object detectors treat object detection as a regression problem, eliminating the need for proposal box generation and reducing computational complexity and runtime. Typical algorithms include SSD\cite{SSD} and YOLO\cite{YOLOV1}. SSD is the first single-stage detector capable of maintaining a certain level of accuracy while ensuring real-time performance. YOLO has performed well in the field of target detection due to its fast processing speed, end-to-end training method, and ability to capture global context information, but its positioning accuracy is low and it performs poorly when faced with category imbalance problems. YOLOv2\cite{YOLO9000} and YOLOv3\cite{YOLOv3} replace GoogLeNet with darknet-19 and darknet-53 as the backbone network, respectively, in addition, global average pooling and Batch Normalization (BN) are adopted to enhance the network performance. YOLOv4~\cite{yolov4} utilizes complete intersection over union (CIOU) loss for prediction box filtering, enhancing model convergence. YOLOv5 incorporates Feature Pyramid Network (FPN) and Pixel Aggregation Network (PAN) structures in its neck network. Subsequent advances in one-stage detection, including RetinaNet\cite{RetinaNet} with focal loss function, fully convolutional one-stage object detector (FCOS)\cite{FCOS}, TOOD\cite{Tood}, which models the object detection task in topological space, and updated version YOLOv8, have improved in terms of speed and accuracy.

\subsection{TIR UAV Target Detection}

Current object detection algorithms are predominantly designed for tasks involving visible light images, with relatively limited research on object detection in TIR images.
SiamSRT\cite{SiamSRT} introduces an innovative Siamese network that is both region-search-free and template-free, using a two-stage architecture. It employs cross-correlation region proposals for initial detection and a similarity-learned region CNN for prediction refinement. This approach integrates spatial location consistency, a temporal repository, and a single-category foreground detector, improving tracking accuracy and robustness against model degradation. However, while effective in stable environments, SiamSRT and similar approaches may struggle in more dynamic or cluttered scenarios, limiting their applicability to more complex TIR detection tasks.
Zou et al.\cite{zou} enhanced the YOLOv5 architecture by incorporating a two-stream backbone network that effectively merges visible light and thermal imaging data. The complementary characteristics of these two modalities are leveraged to improve human detection performance. Additionally, the model integrates a multi-dimensional attention mechanism and a specialized loss function to mitigate background interference and enhance the robustness of inter-modal feature fusion. Despite these improvements, the approach primarily targets pedestrian and vehicle detection and may face challenges in adapting to highly dynamic environments or less common detection tasks, such as wildlife monitoring.

In the domain of TIR-based wildlife detection, CE-RetinaNet \cite{CE-RetinaNet} enhances infrared image feature extraction by integrating a channel enhancement mechanism. It utilizes a batch-normalized random channel attention (BSCA) module to filter out abnormal activations caused by occlusions, ensuring that consistent cross-channel pixels are emphasized. While this improves localization accuracy, the model's reliance on complex channel enhancement and path aggregation operations can result in high computational demands, which may not be feasible for real-time applications or deployment on resource-constrained platforms like UAVs.
Ye et al. \cite{ye} proposed a deep learning algorithm based on the Grid R-CNN framework, specifically addressing the challenge of non-salient target detection in infrared images. The model customizes the feature extraction network to this problem and integrates a guided localization (GA-RPN) mechanism to enhance the region proposal network. By employing a slice inference mechanism, the algorithm efficiently combines multiscale features, resulting in higher-quality proposals and improved target detection accuracy. However, the specialized nature of this approach and its reliance on complex multiscale feature combinations can lead to increased computational overhead, potentially limiting its deployment in real-time or resource-limited settings, such as UAVs.

Our study addresses these challenges by developing a specialized TIR UAV target detection system that achieves high detection accuracy while minimizing both model complexity and computational overhead. This balance between accuracy and efficiency is critical for practical deployment in real-world UAV applications.

\subsection{Small Object Detection}

Existing research on small object detection primarily focuses on multi-scale representation, context information, image super-resolution, and region proposal techniques\cite{Survey_of_Small_Goals}. 

TPH-YOLOv5\cite{TPH-YOLOv5} introduces a small object detection head by replacing the CNN-based prediction head with a Transformer Prediction Head (TPH), and it incorporates a Sparse Localized Attention (SLA) module to efficiently capture asymmetric information between the additional head and other heads, thereby enriching feature representation. However, the complexity introduced by the Transformer structure may lead to increased computational demands, making it less efficient for real-time applications, particularly in scenarios with limited processing power.
Similarly, Zuo et al.\cite{AFFPN} designed the AFFPN architecture, which integrates an Atrous Spatial Pyramid Pooling (ASPP) module into deep feature extraction to capture global contextual information for small targets. Their attention fusion module enhances the spatial and semantic details of multi-level features, improving the accuracy of infrared small target detection. Despite these advancements, the reliance on global contextual information may reduce the model's effectiveness in environments with highly localized features, such as those found in cluttered or noisy backgrounds.
Zhang et al.\cite{CHFNet} proposed the Curvature Half-Level Fusion Network (CHFNet) for single-frame infrared small target detection (IRSTD), which addresses the challenges of dim targets and heavy background clutter by developing a Half-Level Fusion (HLF) block for cross-layer feature fusion. They also introduced a method to calculate the weighted mean curvature of the image to enhance boundary attention and improve edge detection of targets. While this method improves edge detection, the focus on curvature-based features might limit its generalizability to targets with less defined boundaries or those embedded in complex textures.

Moreover, most of these models focus on detecting small objects such as vehicles and pedestrians, while there is limited research on infrared wildlife protection. To address these gaps, our work introduces the LCA module, inspired by Coordinate Attention (CA), which allows the model to capture more fine-grained spatial dependencies and adjust the weights of different feature maps input to the detection head. Additionally, in the construction of the loss function, we emphasize the importance of accurately accounting for the loss of small target bounding boxes, improving the detection capability of small targets through multiple mechanisms. These innovations are particularly effective in complex environments, such as those encountered in infrared wildlife monitoring, where traditional methods have struggled.

\subsection{Lightweight Model Design}
Lightweight network methods include model pruning, network quantization, knowledge distillation, lightweight model design, network compression, and transfer learning. These approaches are designed to minimize the number of model parameters and computational complexity while preserving model performance to suit resource-constrained settings. Given that our research focuses on lightweight model design, the following will focus on the relevant progress in this field.

MobileNet\cite{MobileNets} proposed the concept of depth-separable convolution, decomposing standard convolutions into depth convolutions and point convolutions, thereby significantly reducing the amount of calculation and parameters while preserving high accuracy. SqueezeNet\cite{SqueezeNet} introduces two types of convolutional layers: squeeze layer and expand layer, which achieves lightweight design by reducing the number of channels and increasing the feature map depth. ShuffleNet\cite{ShuffleNet} leverages group convolution and channel shuffling operations to achieve high accuracy with minimal computational cost. ShuffleNetV2\cite{ShufflenetV2} proposes four principles for designing lightweight networks, introduces channel division, and replaces addition operations with connection operations to decrease the number of model parameters.
GhostNet\cite{Ghostnet} introduces the Ghost module to extract redundant features through cost-effective operations, enabling models to effectively utilize and incorporate these features while reducing computational costs.
MobileFormer\cite{Mobile-Former} draws on the design principles of MobileNet and Transformer to achieve a seamless fusion of local features and global features and performs well in a variety of tasks. ConvNext\cite{ConvNet} maximizes the utilization of multi-scale feature information through parallel combination, group convolution, and cross-path design, leading to notable enhancements in efficiency and scalability.

In our work, we introduce the ALSS module, which adheres to the principles of lightweight design, such as bottleneck operation and depthwise convolution strategies, to optimize feature extraction while maintaining a low computational footprint. The ALSS module employs an adaptive channel split strategy and integrates a channel shuffling mechanism, enhancing inter-channel information flow and improving the detection accuracy of blurry targets. This design ensures the model's robustness in handling jitter-induced blur and overlapping targets.
Additionally, we constructed a LCA module that encodes global spatial information, enhancing the network's ability to understand spatial structures. The LCA module combines adaptive pooling and grouped convolutions to effectively capture and integrate spatial information, ensuring high detection precision and robustness against jitter and target overlap while maintaining lightweight operation.

\begin{figure*}[!h]
\centering
\includegraphics[width=15.5cm]{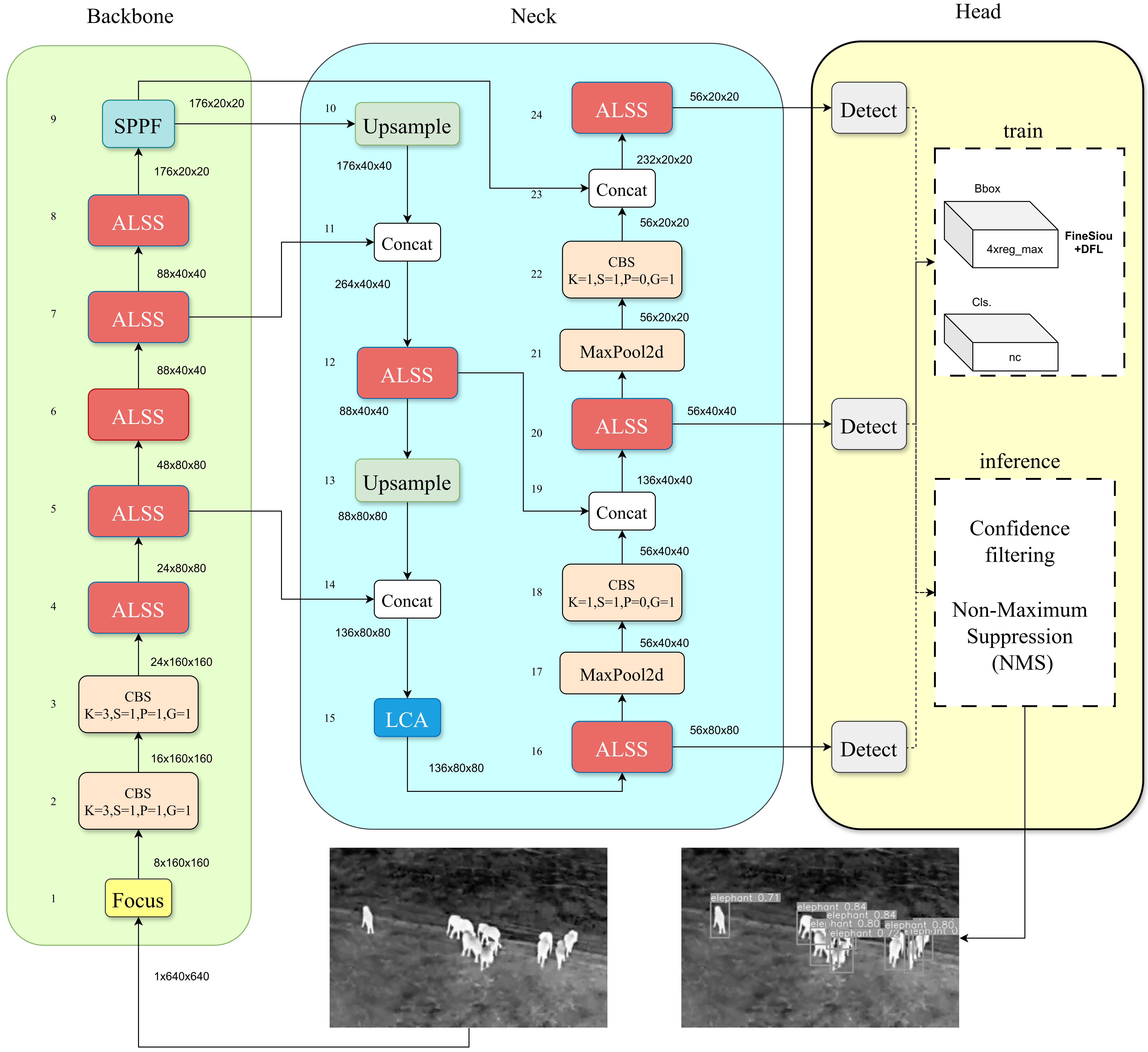}
\caption{The architecture of the ALSS-YOLO detector. CBS denotes Convolution, Batch Normalization, and SiLU activation function. The symbol “k” represents the Kernel size, “s” denotes the Stride, and “p” indicates the Padding. \label{ALSS-YOLO}}
\end{figure*}

\section{Proposed Method}

Our ALSS-YOLO detector's overarching architecture derives from the state-of-the-art YOLOv8-n object detector.
We introduce an innovative ALSS module serving as the backbone network.
By utilizing channel split and subsequent independent processing, the network can learn various features of the input data, improving the model's ability to identify complex or subtle features.
Leveraging bottleneck design and depthwise convolution, the ALSS module achieves efficient feature extraction with a minimal number of parameters. Additionally, through channel shuffling and feature fusion across various paths, the model’s expressiveness and generalization ability are significantly boosted.
Second, inspired by Channel Attention (CA)\cite{CA}, we introduce an LCA module that applies adaptive average pooling to the height and width dimensions of the input feature map, thus generating global contextual features across two spatial dimensions.
Decoupling and independently processing features in each direction, boosts the model's capacity to capture global information, optimizing computational efficiency and parameter usage. 
Besides, A single-channel focus module is specifically designed for single-channel UAV FIR images. This module aggregates the width and height information of a single channel into a four-dimensional channel, thereby retaining more details at a smaller spatial resolution. The width and height data are integrated with channel information via convolution operations, thereby improving the efficiency and accuracy of feature representation.
Finally, FineSIOU was introduced, 
which emphasized the correlation between the shape cost and the dimensions of the ground-truth bounding boxes, and incorporated the angle cost as an independent item into the calculation of the total cost, thus improving the speed of regression and the detection capability for small targets.
Fig.~\ref{ALSS-YOLO} illustrates the architecture of the ALSS-YOLO detector, and Table~\ref{param} outlines the primary parameters.

\begin{table}[!h]
\caption{Parameter count and forward propagation runtime of the main modules in the ALSS-YOLO object detector.\label{param}}
\begin{tabular}{ccccc}
\hline 
\textbf{Number}& \textbf{Module}&\textbf{Output}&\textbf{Params}&\textbf{time(ms)}\\
\hline 
0 	& Input		& 1x640x640		& -			& -		\\
1	& Focus		& 8x160x160		& 1168		& 2.08		\\
2	& CBS		& 16x160x160	& 1184		& 2.08		\\
3	& CBS		& 24x160x160	& 3504		& 1.16		\\
4	& ALSS 	& 24x80x80		& 1728		& 2.29		\\
5	& ALSS 	& 48x80x80		& 3819		& 2.33		\\
6	& ALSS 	& 88x40x40		& 15020		& 2.13		\\
7	& ALSS 	& 88x40x40		& 38393		& 2.12		\\
8	& ALSS 	& 176x20x20		& 54497		& 1.82		\\
9	& SPPF		& 176x20x20		& 77968		& 1.79		\\
10	& Upsample	& 176x40x40		& 0 		& 0.44		\\
11	& Concat	& 264x40x40		& 0 		& 0.05		\\
12	& ALSS 	& 88x40x40		& 379477	& 6.60		\\
13	& Upsample	& 88x80x80		& 0			& 0.10		\\
14	& Concat	& 136x80x80		& 0			& 0.07		\\
15	& LCA		& 136x80x80		& 19448		& 0.81		\\
16	& ALSS 	& 56x80x80		& 77628		& 4.90		\\
17	& MaxPool	& 56x40x40		& 0			& 0.28		\\
18	& CB		& 56x40x40		& 3248		& 0.41		\\
19	& Concat	& 136x40x40		& 0			& 0.05		\\
20	& ALSS 	& 56x40x40		& 120484	& 2.60		\\
21	& MaxPool	& 56x20x20		& 0			& 0.09		\\
22	& CB		& 56x20x20		& 3248		& 0.33		\\
23	& Concat	& 232x20x20		& 0			& 0.05		\\

24	& ALSS 	& 56x20x20		& 263016	& 1.95		\\
25	& Detect	& -				& 391324	& 29.77		\\
-	& Total		& -				& 1455154 *	& 66.28		\\
\hline 
\end{tabular}\\
\noindent{\footnotesize{* The number of parameters after layer fusion is 1452413.}}
\end{table}

\subsection{Lightweight and Efficient Network Architecture - ALSS Module}

\begin{figure}[hptb]
\includegraphics[width=8.5 cm]{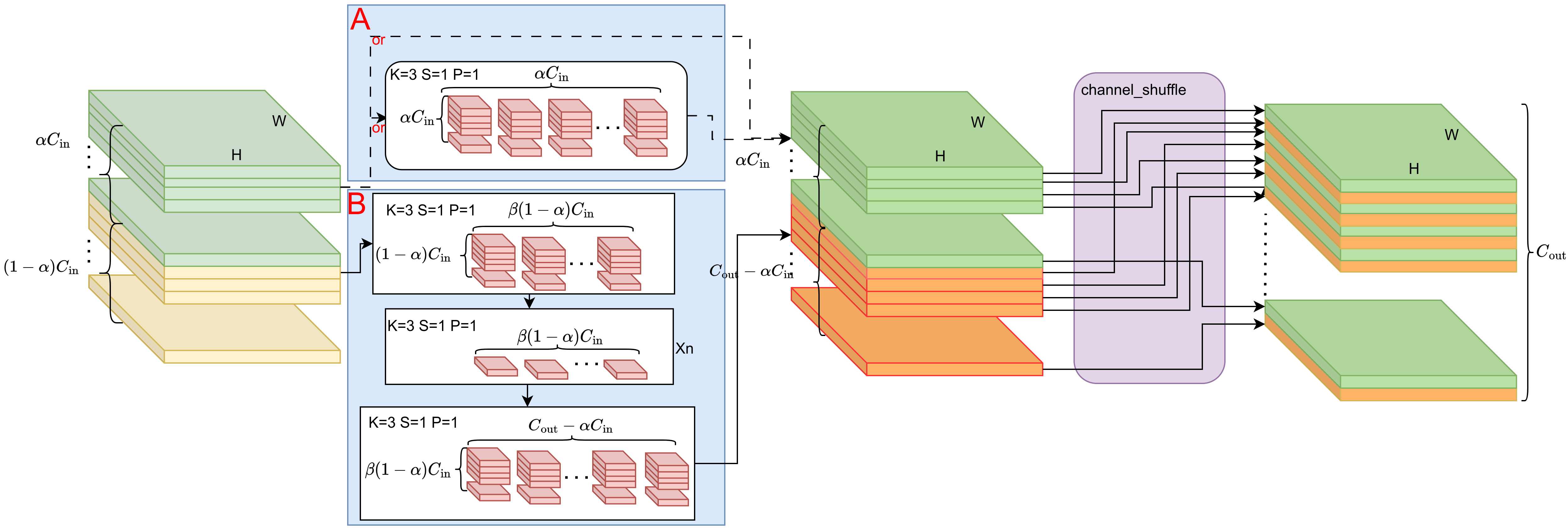}
\caption{ALSS module structure diagram. Part A contains a convolutional layer or identity connection, and part B is a bottleneck structure with depth convolution. All convolutional layers have a step size of 1, and the input and output feature map resolutions are equal.\label{ALSS-S=1}}
\end{figure}

\begin{figure}[hptb]
\includegraphics[width=8.5 cm]{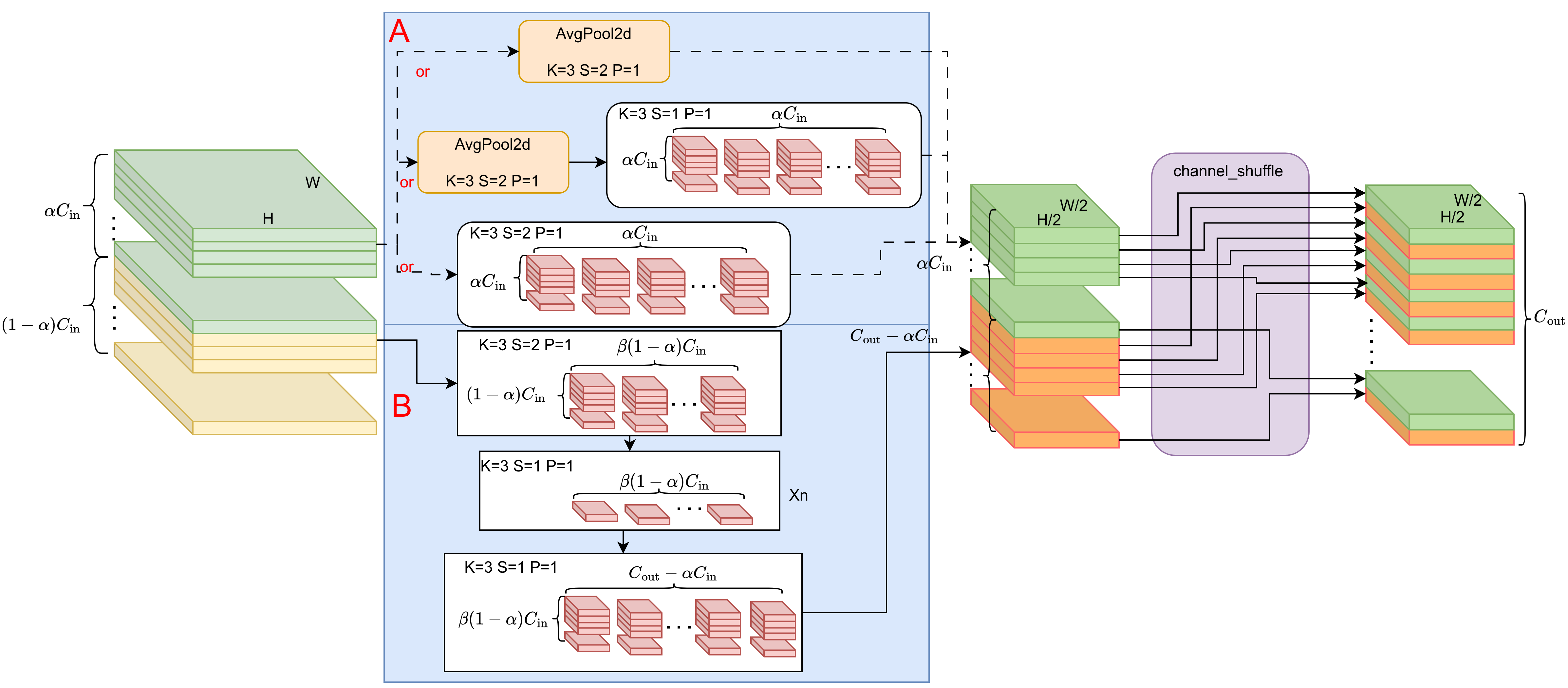}
\caption{ALSS module structure diagram.
Part A contains max pooling with stride 2, or max pooling with stride 2 concatenating convolutions with stride 1, or convolutions with stride 2, part B is a bottleneck structure with depth convolution, in which the first convolution of the bottleneck structure uses a convolution with a stride of 2. The width and height of the output feature map are half those of the input feature map, and the model has a downsampling effect. 
\label{ALSS-S=2}}
\end{figure}

In deep learning, lightweight and efficiency is one of the current research hotspots. Designing a lightweight and efficient neural network architecture is crucial for scenes with limited resources. In this context, we introduce a novel network architecture named ALSS, aimed at achieving lightweight and efficient object detection. The module's structure is illustrated in Fig.~\ref{ALSS-S=1}. Additionally, Fig.~\ref{ALSS-S=2} showcases the configuration of the ALSS module when serving as a downsampling operation. 

This module adopts a narrow and deep design strategy, first splitting the input feature map $X \in \mathbb{R}^{H \times W \times C}$ into $X_{A\text{in}} \in \mathbb{R}^{H \times W \times \alpha C}$ and $X_{B\text{in}} \in \mathbb{R}^{H \times W \times (1-\alpha) C}$:
\begin{equation}
\begin{split}
X_{A\text{in}}, X_{B\text{in}} = \text{Split}(X \in \mathbb{R}^{H \times W \times C})
\end{split}
\end{equation}
where $C$ represents the total number of input channels.
The scaling factor \(\alpha\)\ is dynamically adjusted according to the feature level within the ALSS module to meet the varying requirements of feature abstraction at different stages of the network.
At lower feature levels, where the focus is on capturing fine-grained details and low-level features, we set a low $\alpha$ value. This ensures that only a small portion of the channels is processed through part A of Fig.~\ref{ALSS-S=1}, which typically involves basic convolutional operations. The majority of the channels are directed into a more complex multi-level network (part B of Fig.~\ref{ALSS-S=1}), designed for enhanced feature extraction and capturing intricate patterns.
As the feature level increases, the need for abstract and high-order semantic information becomes more critical. To accommodate this, we increase the \(\alpha\) value at higher feature levels. This adjustment connects the input channels directly to the output channels, facilitating the extraction of complex features and enhancing the network's representation capacity. By prioritizing convolutional operations at lower levels and shifting towards a ResNet-like structure for residual learning at higher levels, this strategy balances the trade-off between computational efficiency and feature richness, optimizing the network's performance across different stages of the feature hierarchy.

Furthermore, at lower feature levels, in the network structure shown in part A of Fig.~\ref{ALSS-S=1}, we use convolution operations to extract features. This is because, in shallow convolutional layers, features are more localized and detailed. Through convolutional operations, the network can acquire a diverse range of filters to detect local features in the image, including edges, textures, and other characteristics. This methodology empowers the network to efficiently extract basic features and progressively amalgamate them into more advanced feature representations. Conversely, at higher feature levels, we prefer employing identity connections. In deep convolutional layers, features become more abstract and semantic, containing higher-level semantic information such as object shapes, categories, etc. By utilizing identity connections, we can preserve these high-level features and prevent excessive compression or loss of important information learned by the network. This process can be expressed as:
%

\begin{flalign}
X_{A\text{out}}& \in \mathbb{R}^{H \times W \times \alpha C} &
\end{flalign}
\begin{flalign*}
& = 
\begin{cases} 
\text{Conv}\left(X_{A\text{in}} \in \mathbb{R}^{H \times W \times \alpha C}\right), & \text{for lower layers} \\[10pt]
X_{A\text{in}}\in \mathbb{R}^{H \times W \times \alpha C}, & \text{for higher layers}
\end{cases} &
\end{flalign*}
This top-down information propagation facilitates the flow and sharing of information, ultimately boosting the model's performance and generalization capabilities. The rationality of the parameter setting strategy for \(\alpha\) and the network structure selection strategy for Part A will be validated in the subsequent experimental section.

Intending to diminish the model's parameter count and enhance computational efficiency, we use a bottleneck structure and a depthwise convolution (part B of Fig.~\ref{ALSS-S=1}). The bottleneck structure effectively reduces computational costs by extracting features at a lower dimension, while the depthwise convolution further improves the network's perception range and feature extraction capabilities, and is particularly suitable for processing more complex semantic information.
Specifically, a $3 \times 3 $ convolutional kernel is initially applied with a step size of 1 to accomplish dimensionality reduction. 
\begin{flalign}
X_{B_\text{Phase 1}} & \in \mathbb{R}^{H \times W \times \beta(1-\alpha)C} &
\end{flalign}
\begin{flalign*}
& = \text{Conv}\left(X_{B\text{in}} \in \mathbb{R}^{H \times W \times (1-\alpha)C}\right) &
\end{flalign*}

This operation introduces a dimensionality reduction coefficient \(\beta\), which modifies the number of output channels to regulate the model's parameter count and computational workload. 
In practice, when processing levels with a higher number of channels, a larger \(\beta\) value is chosen to significantly reduce resource consumption by compressing the channels more aggressively. This allows the model to manage the increased computational load and memory usage effectively. Conversely, at levels with lower channel numbers, smaller \(\beta\) values are employed to ensure that the model can maintain sufficient feature extraction capabilities, preserving critical information in the feature maps. The choice of \(\beta\) is therefore a balance between resource efficiency and the ability to capture and represent essential features across different stages of the network. The use of depthwise convolutions further complements this strategy by reducing the number of parameters and computational complexity while maintaining spatial and channel-wise relationships in the data, contributing to a lightweight yet effective architecture.

After the dimensionality reduction operation, To further augment the model's feature extraction capability, this strategy then applies a layer of $3 \times 3 $ depthwise convolution. Depthwise convolution enhances the network's nonlinearity by conducting convolution on each channel of the input feature map individually, thus decreasing the parameter count as opposed to traditional convolution;
\begin{flalign}
X_{B_\text{Phase 2}} & \in \mathbb{R}^{H \times W \times \beta(1-\alpha)C} &
\end{flalign}
\begin{flalign*}
& = \text{DWConv}^n(X_{B_\text{Phase 1}} \in \mathbb{R}^{H \times W \times \beta(1-\alpha)C}) &
\end{flalign*}
Then, To adjust the output dimension of the model, a $3 \times 3 $ convolution kernel is applied again, and the output dimension is set to \(C_{\text{out}}-\alpha C\), 
\begin{flalign}
X_{B_\text{out}} & \in \mathbb{R}^{H \times W \times (C_{\text{out}}-\alpha C)} &
\end{flalign}
\begin{flalign*}
& = \text{Conv}\left(X_{B\text{Phase 2}} \in \mathbb{R}^{H \times W \times \beta(1-\alpha)C}\right) &
\end{flalign*}
where \(C_{\text{out}}\) represent the number of output channels, and \(\alpha\) denotes the predefined scaling factors mentioned previously.

Our network design also follows the lightweight network design principles mentioned in ShuffleNetV2\cite{ShufflenetV2}, avoiding excessive branch structures and choosing to concatenate feature maps along the channel direction to reduce the computational burden.
This design not only enhances the network's computational efficiency but also improves the information exchange and feature representation capabilities, making the network more suitable for practical applications in various complex scenarios. Following the advice of Ref.\cite{Depthwise}, the SiLU activation function is not employed following the depthwise convolution.

In the final stage of the module, channel shuffling operations were employed to enhance information exchange among different feature channels. This rearranges the input feature map channels, enabling previously isolated feature branches to share learned information, thereby improving information flow and feature expression diversity. The entire network architecture, while maintaining a lightweight design, better captures and integrates multi-scale, multi-angle feature information, enhancing the capability to extract features from complex scenes.

In the ALSS module, which acts as a downsampling structure, we designed three modules to downsample the \(\alpha C_{\text{in}}\)
input feature map channels. As shown in part A of Fig.~\ref{ALSS-S=2}, according to the feature level within the network model, we apply the following operations in ascending order from low to high: convolution, convolution-pooling, and pooling;
\begin{flalign}
X_{A\text{out}}& \in \mathbb{R}^{H/2 \times W/2 \times \alpha C} &
\end{flalign}
\begin{flalign*}
& = 
\begin{cases} 
\text{Conv}\left(X_{A\text{in}} \in \mathbb{R}^{H \times W \times \alpha C}\right), & \text{for lower layers} \\[10pt]
\text{Conv}(\text{Pool}(X_{A\text{in}} \in \mathbb{R}^{H \times W \times \alpha C})), & \text{for middle layers} \\[10pt]
\text{Pool}(X_{A\text{in}} \in \mathbb{R}^{H \times W \times \alpha C}), & \text{for higher layers}
\end{cases} &
\end{flalign*}

These operations can achieve detailed feature extraction at low levels and help capture the basic texture and shape information of the image. At higher levels, a residual learning structure akin to ResNet is incorporated to facilitate effective training of the network, particularly in cases where it is exceptionally deep. This method enables the network to better learn high-level abstract features without losing low-level information. For the \((1-\alpha)C_{\text{in}}\) input feature map channel, a stride of 2 is utilized in the initial $3 \times 3 $ convolution within the bottleneck stage to decrease the width and height of the feature map;
\begin{flalign}
X_{B_\text{Phase 1}} & \in \mathbb{R}^{H/2 \times W/2 \times \beta(1-\alpha)C} &
\end{flalign}
\begin{flalign*}
& = \text{Conv}\left(X_{B\text{in}} \in \mathbb{R}^{H \times W \times (1-\alpha)C}\right) &
\end{flalign*}

Within the ALSS-YOLO network architecture, the $\alpha$ coefficients from bottom to top are set as (0.4, 0.4, 0.5, 0.6, 0.7, 0.2, 0.3, 0.2, 0.2), while the $\beta$ coefficients are defined as (0.4, 0.4, 0.5, 0.6, 0.6, 0.8, 0.8, 0.8, 0.8).

\subsection{LCA module}
\begin{figure}[hptb]
\includegraphics[width=8.5 cm]{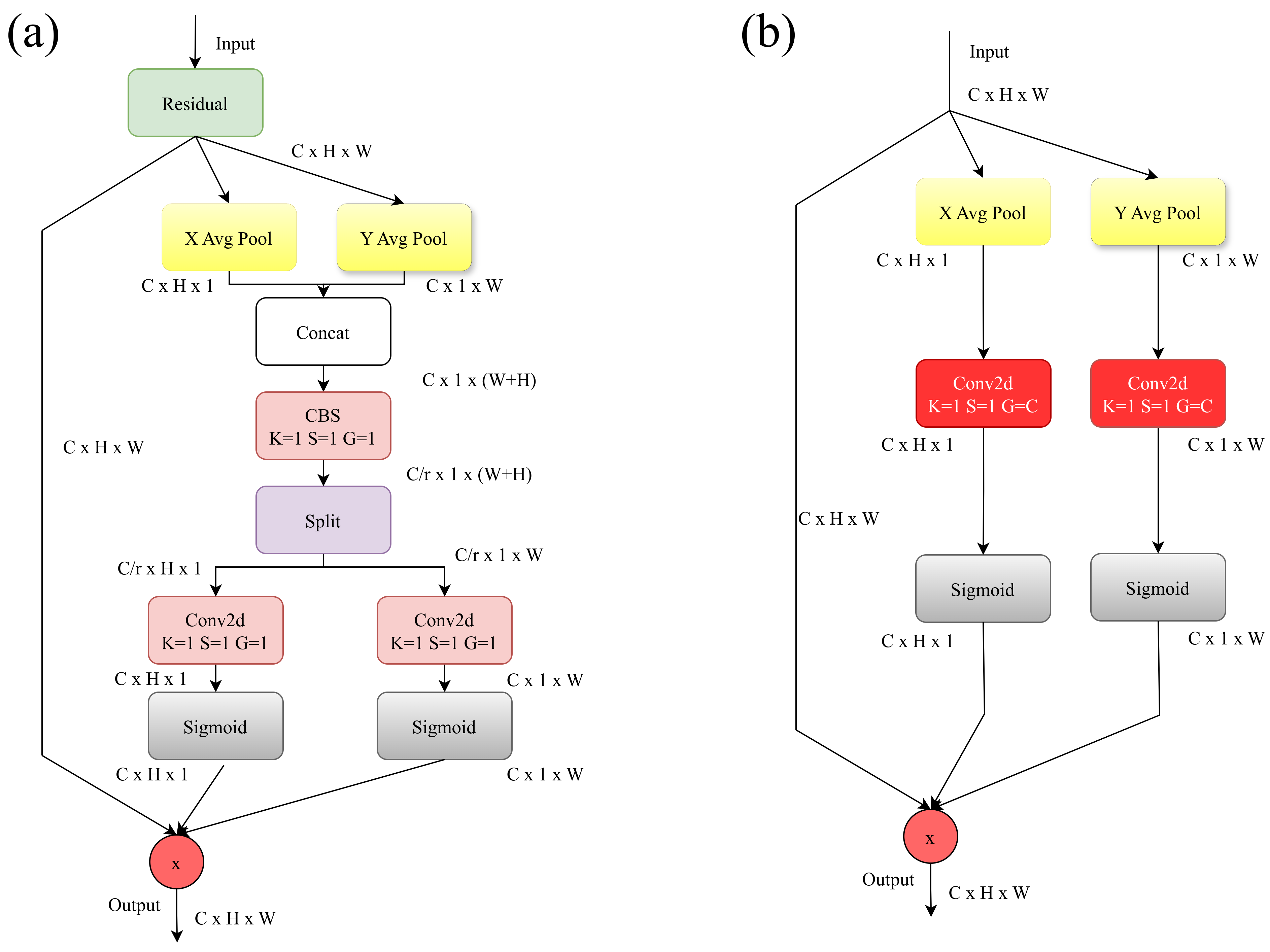}
\caption{Schematic diagram of the structure of the CA and LCA modules: (\textbf{a}) CA module. (\textbf{b}) LCA module.\label{LCA2}}
\end{figure}

\begin{figure}[hptb]
\includegraphics[width=8.5 cm]{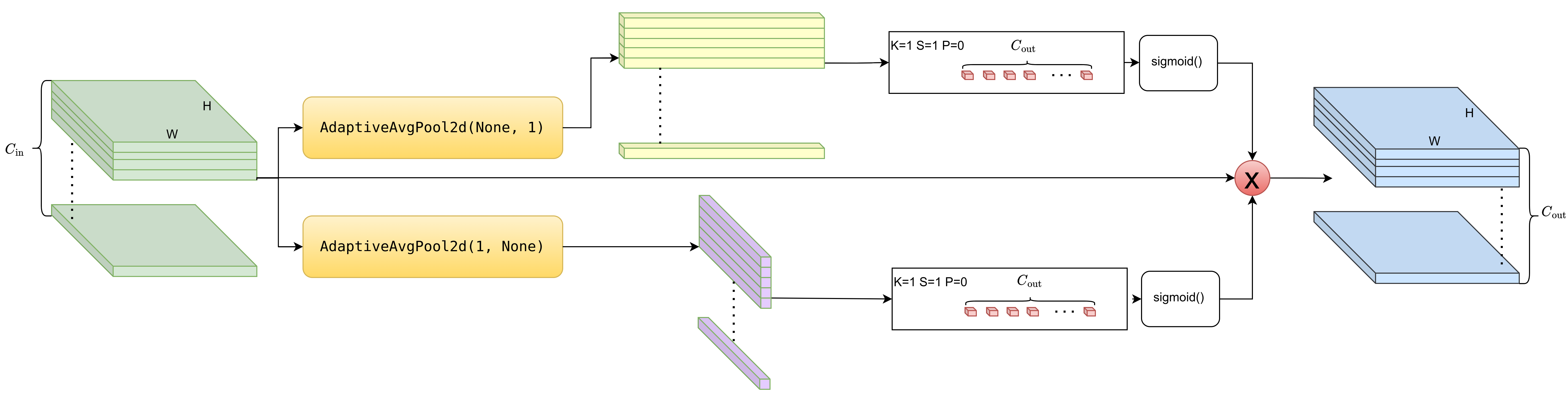}
\caption{Detailed internal structure of the LCA model.\label{LCA}}
\end{figure}

Recent advancements in deep learning have highlighted the significance of attention mechanisms in enhancing neural network performance across various tasks. Specifically, attention modules have been instrumental in refining feature representations by selectively emphasizing informative components and suppressing irrelevant ones within feature maps.
In particular, the introduction of the CA\cite{Coordinate_attention} mechanism further optimizes this process. It not only pays attention to the channel dimension of the feature but also emphasizes the coordinate information of the feature space. This design enables the model to more effectively integrate global context information while parsing the spatial dimension.
To enhance the efficiency of feature extraction and minimize the number of parameters, this paper proposes an improved coordinate attention mechanism, namely LCA. Fig.~\ref{LCA2} provides a comparative view between CA and LCA, while Fig.~\ref{LCA} reveals the internal structure of LCA in detail. In the following sections, we will elaborate on the design principles and implementation details of LCA.

For the input tensor $X$ with dimensions $C \times W \times H$, where $H$, $W$, and $C$ represent height, width, and the number of channels, respectively, we employ a specialized approach to capture features along specific spatial dimensions. This is achieved through the application of two independent pooling operations, each utilizing a distinct kernel size: one uses a pooling kernel of $(H, 1)$ along the horizontal dimension, while the other employs a pooling kernel of $(1, W)$ along the vertical dimension. These operations are performed separately on each channel to preserve spatial information along the respective dimensions.
For the $c$-th channel in the input tensor $X$, horizontal pooling with a kernel size of $(H, 1)$ is applied at height $h$, and vertical pooling with a kernel size of $(1, W)$ is applied at width $w$. The pooled output $z_{c}^{h}(h)$
at position $(h, c)$ is computed as:

\begin{equation}
z_{c}^{h}(h)=\frac{1}{W} \sum_{0 \leq i<W} x_{c}(h, i)
\end{equation}

and the pooling output $z_{c}^{w}(w)$ at position $(w, c)$\cite{Coordinate_attention}:
\begin{equation}
z_{c}^{w}(w)=\frac{1}{H} \sum_{0 \leq j<H} x_{c}(j, w)
\end{equation}

These two directional pooling operations enable the capture of features along both spatial dimensions, resulting in a direction-aware feature map that enhances the network's ability to recognize precise locations and improve spatial awareness. This novel method of decoupling feature processing along both dimensions allows for efficient and cost-effective feature extraction, particularly advantageous for mobile devices with limited computational resources.

Once the feature maps are generated through the aforementioned pooling operations, they are subjected to distinct $1 \times 1$ depthwise separable convolutions. These convolutions facilitate linear transformations between channels while incorporating nonlinear activation functions to modulate attention weights. The resulting transformations are expressed as follows:
\begin{equation}
\label{h}
g^{h}=\sigma\left(F_{h}\left(z_{c}^{h}(h)\right)\right.
\end{equation}
\begin{equation}
\label{w}
g^{w}=\sigma\left(F_{w}\left(z_{c}^{w}(w)\right)\right.
\end{equation}
here, $\sigma$ represents the nonlinear activation function, $g^{h}$ and $g^{w}$ denote the adjusted attention weights. The final output of our coordinate attention block can thus be expressed as:
\begin{equation}
y_{c}(i, j)=x_{c}(i, j) \times g_{c}^{h}(i) \times g_{c}^{w}(j)
\end{equation}

The LCA module further refines the network's ability to focus on spatial features. By applying adaptive average pooling along the height and width dimensions, followed by depthwise separable convolutions, the module captures global spatial information while maintaining computational efficiency. The attention weights generated through these processes modulate the input tensor, enhancing the network's discriminative capabilities.

When integrating the LCA module into the network architecture, particular attention is given to layers with higher-resolution feature maps, especially for detecting small objects. Based on our analysis, small objects are more effectively detected on these feature maps. Therefore, the LCA module is strategically placed at the 15th layer, which is close to the detection head, to maximize the detection accuracy for small objects in complex scenarios. We will also verify its rationality in the following experimental section.

In conclusion, the LCA module's combination of adaptive pooling and depthwise separable convolutions presents a robust mechanism for enhancing spatial attention while maintaining low computational complexity. Its strategic integration into the network architecture significantly improves feature discrimination and detection capabilities, particularly in environments characterized by complex spatial structures and the presence of small, overlapping objects, making it an essential component for tasks requiring precision and efficiency.

\subsection{Single Channel Focus module}
\begin{figure}[hptb]
\includegraphics[width=8.5 cm]{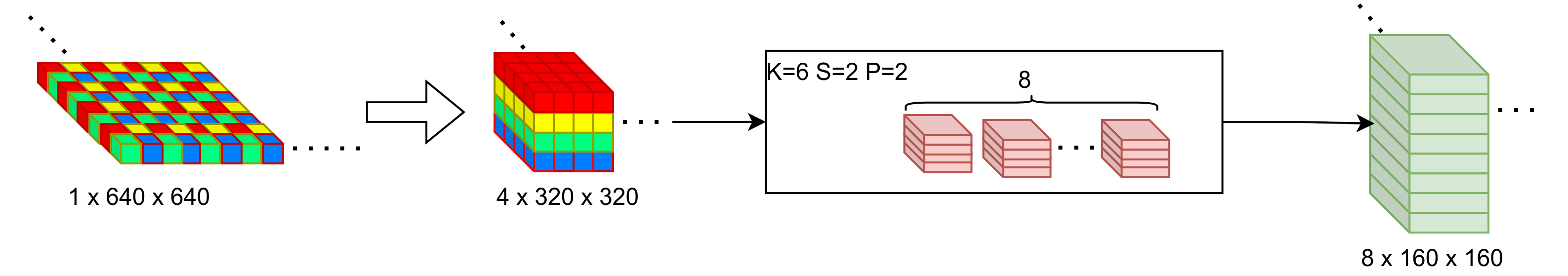}
\caption{Single-channel focus module structure diagram.\label{Focus}}
\end{figure}

An efficient and accurate feature extraction mechanism can substantially enhance the model's performance. For single-channel TIR images, traditional feature extraction methods face the challenge of insufficient information utilization or high consumption of computing resources. To this end, we propose a single-channel focus module, which aims to boost the model's feature representation capabilities by efficiently aggregating width and height information into the channel dimension and through an optimized convolution strategy(refer to Fig.~\ref{Focus}).

For a given $x \in \mathbb{R}^{N \times 1 \times H \times W}$  TIR image, where $N$ is the number of samples (batch size), $1$ is the number of channels, \( H \) and \( W \) are the height and width respectively.
It is first segmented in the width and height dimensions, and the segmented areas are combined in the channel dimension to generate a $x \in \mathbb{R}^{N \times 4 \times H/2 \times W/2}$ image, as shown in Eq.\eqref{focus}. Each channel carries the local information in the original TIR image, and the new multi-channel combination fuses this local information in the channel dimension, thereby achieving the effect of information aggregation.
\begin{equation}
\label{focus}
\begin{split}
x_{00} = x[:, 0, 0:H:2, 0:W:2]\\ x_{01} = x[:, 0, 0:H:2, 1:W:2]\\ 
x_{10} = x[:, 0, 1:H:2, 0:W:2]\\ x_{11} = x[:, 0, 1:H:2, 1:W:2]
\end{split}
\end{equation}
\begin{equation}
x_{\text{concat}} = \text{concat}(x_{00}, x_{01}, x_{10}, x_{11}, \text{dim}=1)
\end{equation}
For $x_{00}$, it denotes the selection of pixels starting from position (0, 0) with a spacing of every 2 pixels; while $x_{01}$, $x_{10}$, $x_{11}$ follow the same pattern.
Subsequently, the generated feature map is processed by a convolution operation with a convolution kernel size of $6$ and a stride of $2$(refer to Eq.\eqref{focus2}), which expands the receptive field for global feature extraction, aiding in capturing spatial information more effectively from the input feature map. This further helps the model learn more global and abstract features, improves the ability to recognize complex patterns and structures, and halves the size of the feature map. This approach helps to decrease the number of parameters and computational load in subsequent layers while preserving key feature information.
\begin{equation}
\label{focus2}
y = \text{conv}_{k=6,s=2}(x_\text{concat})
\end{equation}

\subsection{Optimization of Loss Function - FineSIOU}
\begin{figure}[hptb]
\includegraphics[width=6.2 cm]{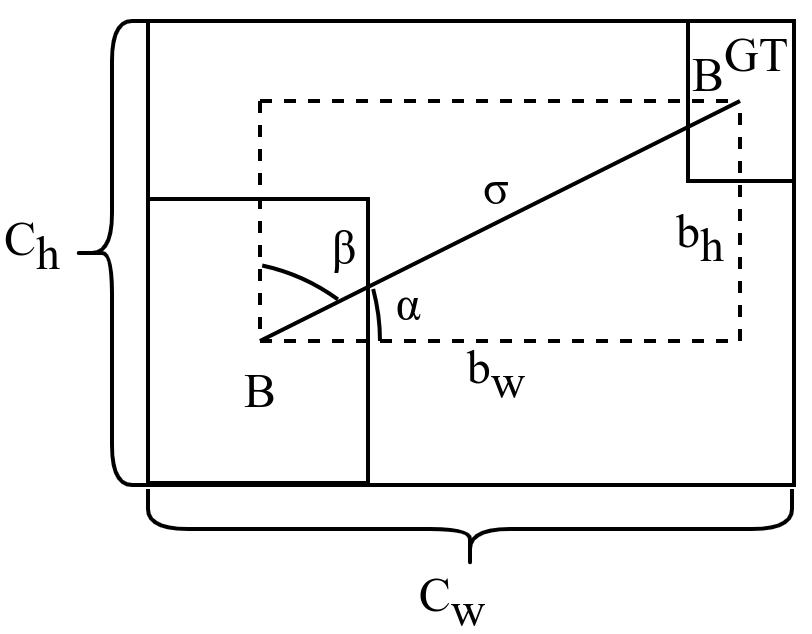}
\caption{Schematic diagram of SIOU calculation scheme\label{SIOU.png}}
\end{figure}

In YOLOv8, the loss function comprises classification loss and bounding box regression loss. Compared to the CIOU used in YOLOv8 for bounding box regression loss, the SIOU\cite{SIOU} loss function takes into account the angle between the true box and the predicted box. It makes the model easier and faster to approach the ground-truth box during training, resulting in a notable enhancement of the model's training efficiency and accuracy, especially when detecting objects in complex backgrounds and overlapping scenarios. Building upon the SIOU, we consider the close relationship between bounding box regression and the size of the ground-truth box (regression difficulty being greater for small targets than for large targets) and improve upon SIOU by adjusting the shape loss to emphasize the size and shape of the ground-truth box, notably enhancing the accuracy of small object detection. Additionally, the angle cost is separated as an individual term to accelerate convergence. This enhanced loss function is referred to as FineSIOU. To provide context, it is essential to briefly review SIOU, which integrates four components: angle cost, distance cost, shape cost, and IOU cost. Fig.~\ref{SIOU.png} shows the SIOU calculation scheme.

SIOU defines Shape cost as\cite{SIOU}:
\begin{equation}
\Omega=\sum_{t=w, h}\left(1-e^{-\omega_{t}}\right)^{\theta}
\end{equation}
\begin{equation}
\omega_{w}=\frac{\left|w-w^{g t}\right|}{\max \left(w, w^{g t}\right)}, \omega_{h}=\frac{\left|h-h^{g t}\right|}{\max \left(h, h^{g t}\right)}
\end{equation}
$w^{g t}$ and $h^{g t}$ represent the width and height of the ground truth box, respectively. $W$ and $h$ represent the width and height of the predicted box, respectively. $\theta$ is used to control the degree of attention to shape loss. 

The loss function of SIOU is defined as\cite{SIOU}:
\begin{equation}
	L_{b o x}=1-I O U+\frac{\Delta+\Omega}{2}
\end{equation}
$I O U$ represents the IoU loss, $\Delta$ denotes the distance cost, $\Omega$ corresponds to the shape cost, and the angle cost $\lambda$ is integrated into the calculation of the distance cost.

\begin{figure}[hptb]
\includegraphics[width=8.5 cm]{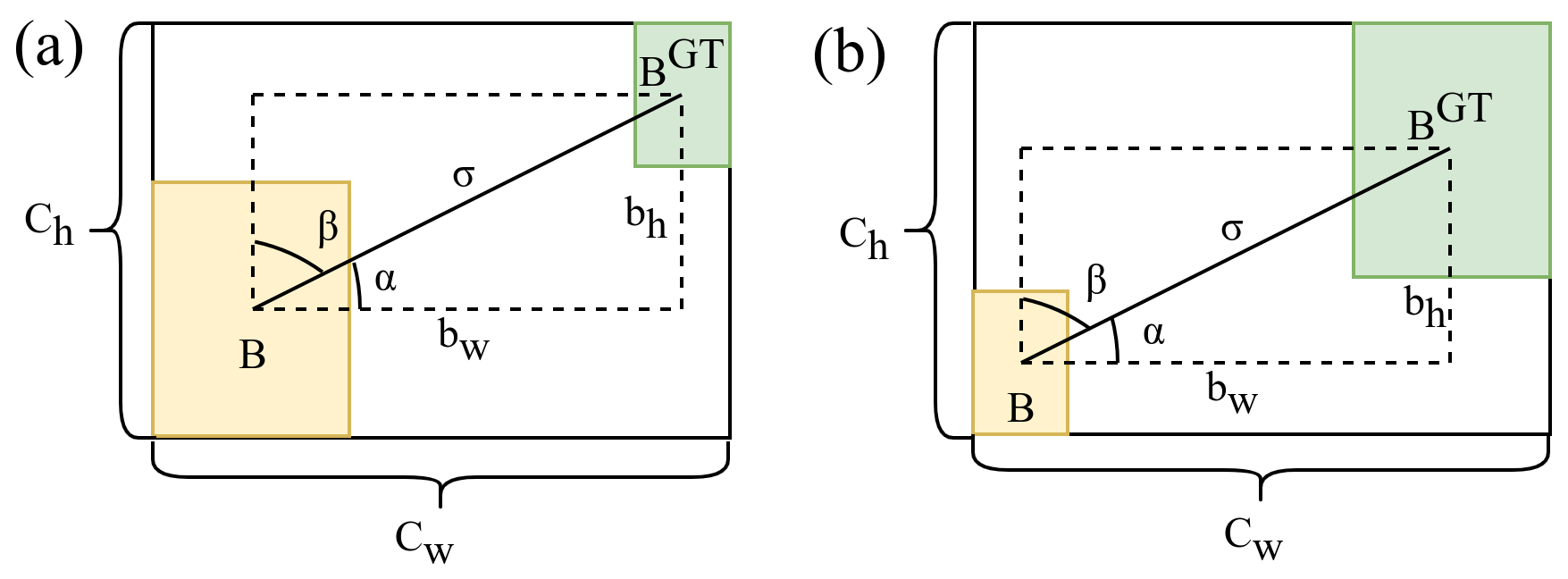}
\caption{Given a fixed angle cost and distance cost, the shape cost shown in (\textbf{a}) should be higher than that in (\textbf{b}).\label{siou1.png}}
\end{figure}

\begin{figure}[hptb]
\includegraphics[width=8.5 cm]{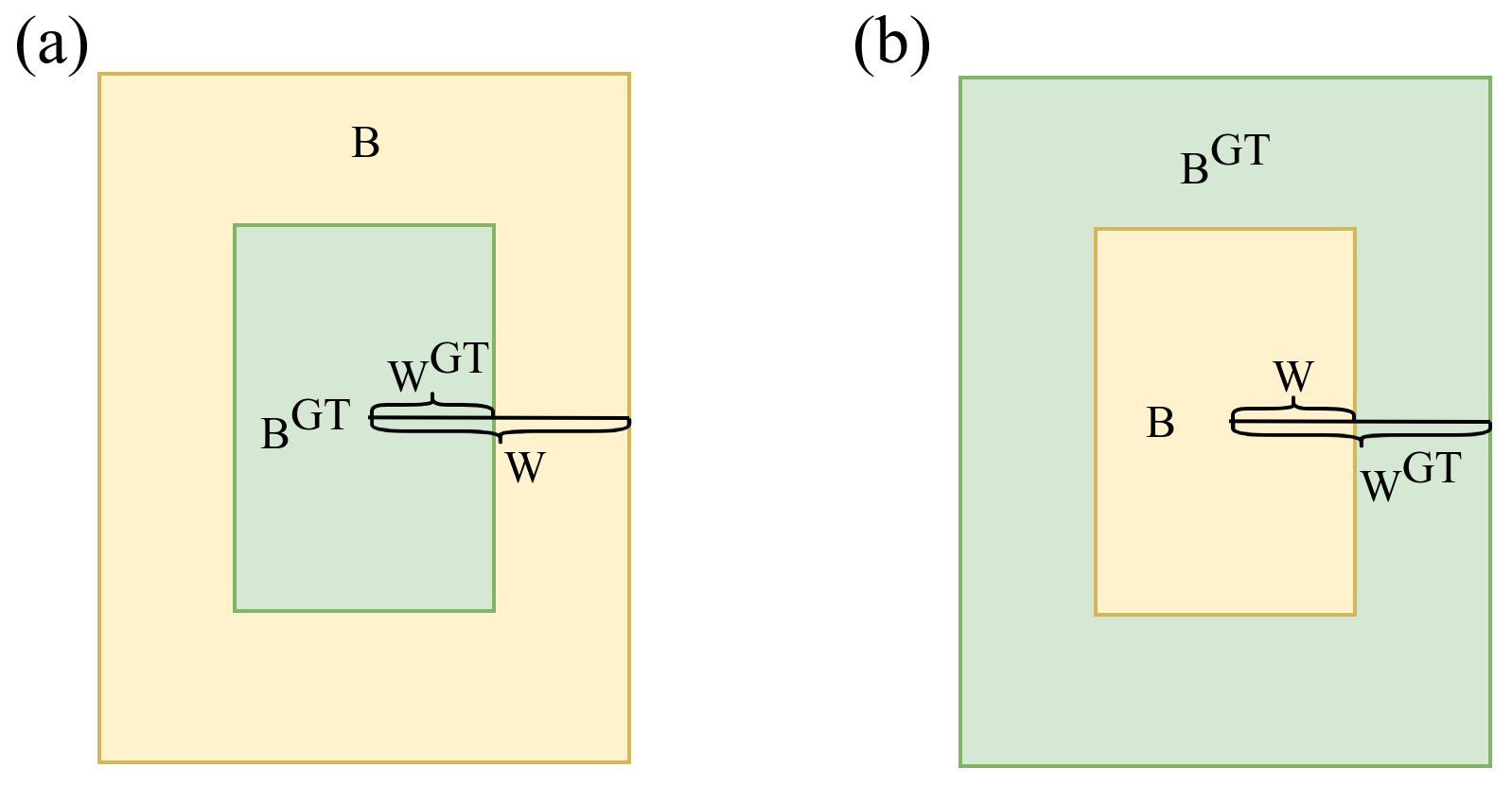}
\caption{For the specific case where the distance cost is zero, the shape cost shown in (\textbf{a}) should be greater than (\textbf{b}).\label{siou2.png}}
\end{figure}

In our investigation into the calculation of shape costs, we observed that the difficulty of shape regression is closely related to the size of the ground-truth box. As shown in Fig.~\ref{siou1.png}(a,b), under the premise of keeping the angle cost and distance cost unchanged, the shape cost results obtained by SIOU are the same. In contrast, we believe that the regression difficulty between the small ground-truth box and the large predicted box in Fig.~\ref{siou1.png}(a) should be higher than that of the large ground-truth box to the small predicted box in Fig.~\ref{siou1.png}(b). Furthermore, in the specific case where the distance cost is zero as shown in Fig.~\ref{siou2.png}(a,b), we believe that the shape cost of Fig.~\ref{siou2.png}(a) should be lower than that of Fig.~\ref{siou2.png}(b). Based on the above analysis, the shape cost of FineSIOU we devised is as follows:
\begin{equation}
	\Omega=\sum_{t=w, h}\left(1-e^{-\omega_{t}}\right)^{\theta}
\end{equation}
\begin{equation}
	\omega_{w}=\frac{\left|w-w^{g t}\right|}{w^{g t}}, \omega_{h}=\frac{\left|h-h^{g t}\right|}{h^{g t}}
\end{equation}
The parameter $\theta$ is utilized to modulate the level of emphasis on the shape loss. When $\theta$ is assigned a value of 1, the shape optimization process will take priority. Our experimental results show that setting it to 6 achieves good results. The refined shape cost pays more attention to the size of the ground-truth box, especially for aerial images dominated by small objects. Smaller bounding boxes will produce higher shape loss values, which is consistent with the intuitive view that smaller objects pose greater regression challenges than larger objects. Table~\ref{FineSIOU_example} provides examples illustrating the shape cost computed under various ground truth and predicted bounding boxes.

\begin{table*}[!t]
\centering
  \caption{Parameter count and forward propagation runtime of the main modules in the ALSS-YOLO object detector.\label{FineSIOU_example}}
  \begin{tabular}{ccccccccccc}
      \hline 
      \textbf{Examples}& \textbf{$w^{g t}$}&\textbf{$h^{g t}$}&\textbf{$w$}&\textbf{$h$}&\textbf{$\omega_{w}$}(SIOU)&\textbf{$\omega_{w}$}(FineSIOU)&\textbf{$\omega_{h}$}(SIOU)&\textbf{$\omega_{h}$}(FineSIOU)&\textbf{$\Omega$}(SIOU)&\textbf{$\Omega$}(FineSIOU)\\
      \hline 
      \multirow{4}{*}{0} 	& 30	& 40	& 50	& 60	& 2/5	& 2/3	          & 1/3	& 1/2	& 0.0586	& \textbf{0.1761}	\\
                  & 30	& 40	& 10	& 20	& 2/3	& 2/3	& 1/2	& 1/2	& 0.0967	& \textbf{0.1761}	\\
                  & 40	& 30	& 60	& 50	& 1/3	& 1/2	& 2/5	& 2/3	& 0.0586	& \textbf{0.1761}	\\
                  & 50	& 60	& 30	& 40	& 2/5	& 2/5	& 1/3	& 1/3	& 0.0586	& \textbf{0.0586}	\\

      \hline
      \multirow{4}{*}{1}	& 60	& 80	& 80	& 100	& 1/4	& 1/3	          & 1/5	& 1/4	& 0.0168	& \textbf{0.0336}	\\
                  & 60	& 80	& 40	& 60	& 1/3	& 1/3	& 1/4	& 1/4	& 0.0336	& \textbf{0.0336}	\\
                  & 80	& 60	& 100	& 80	& 1/5	& 1/4	& 1/4	& 1/3	& 0.0167	& \textbf{0.0336}	\\
                  & 80	& 100	& 60	& 80	& 1/4	& 1/4	& 1/5	& 1/5 	& 0.0167	& \textbf{0.0168}	\\
      \hline
    \end{tabular}
\end{table*}

At the same time, we also focus on the angle cost and incorporate it into the calculation of the total cost as a separate item. Our modified angle cost is:
\begin{equation}
\Lambda=1-2 \times \sin ^{2}\left(\arcsin \left(\frac{b_{h}}{\sigma}\right)-\frac{\pi}{4}\right)
\end{equation}

\begin{equation}
	\zeta = \left(1-e^{-\left(0.9847 - \Lambda\right)}\right)^{\eta }
\end{equation}
Here,$\sigma$ represents the Euclidean distance between the center point of the ground-truth box and the predicted box, while $b_{h}$ indicates the height difference between the center point of the ground-truth box and the predicted box.
The calculated value of 0.9847 corresponds to an angle of $\alpha$=5°. When the angle is less than 5°, the angle loss between the predicted box and the ground-truth box is no longer the main focus, but the distance loss and shape loss should be paid more attention. $\eta$ is a parameter utilized to regulate the level of emphasis on the angle loss, In our experiments, setting it to 3 yielded favorable outcomes, we posit that finer parameter tuning may lead to even better results. Finally, FineSIOU is expressed as:
\begin{equation}
L_{b o x}=1-I O U+\frac{\zeta+\Delta+\Omega}{2}
\end{equation}

\section{Experiment and Analysis}
\subsection{Experimental Setup and Evaluation}

\subsubsection{Experimental data}

In this research, we employed the TIR UVA dataset BIRDSAI\cite{BIRDSAI}, which was made public by Harvard University in 2020. This dataset is a comprehensive collection obtained from TIR cameras mounted on fixed-wing drones, capturing nocturnal events of animals and humans in protected regions across various African countries. It includes challenging scenarios such as orientation alterations and background interference due to heat reflections, numerous camera rotations, and motion blurring. The dataset comprises 48 authentic aerial TIR videos of varying durations, with meticulous annotations of objects such as humans and wildlife, along with their trajectories.
The original dataset contains a total of 9 categories, but since the proportion of the other categories in the entire dataset is small, we selected
four genes for experimental research, namely unknown, human, elephant, and lion, and only selected images with labels. Our training set consisted of 10,924 images, while the test set included 1,943 images. Rather than applying offline data augmentation to the original images, we implemented a series of online augmentation techniques such as mosaic augmentation, random perturbations (including rotation and scaling), mixup, color perturbation, and random flipping.

\begin{figure}[hptb]
\includegraphics[width=8 cm]{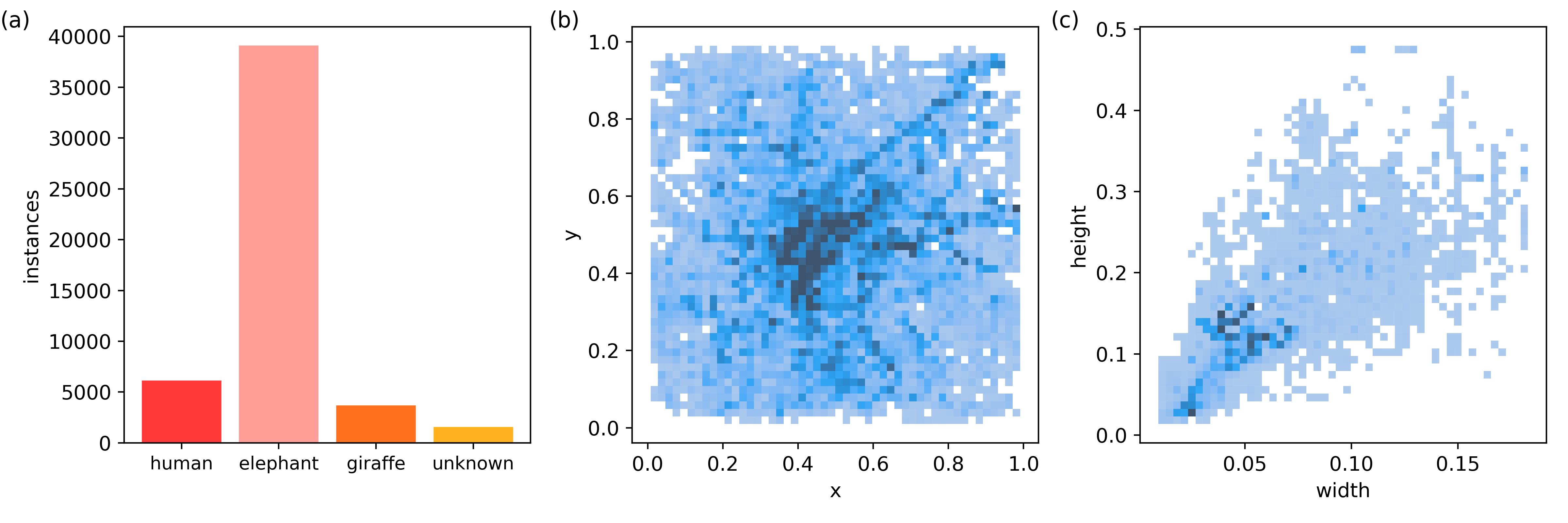}
\caption{Label distribution of the ground-truth boxes in the train set.\label{combined_plots.jpg}}
\end{figure}

\begin{figure}[hptb]
\includegraphics[width=8 cm]{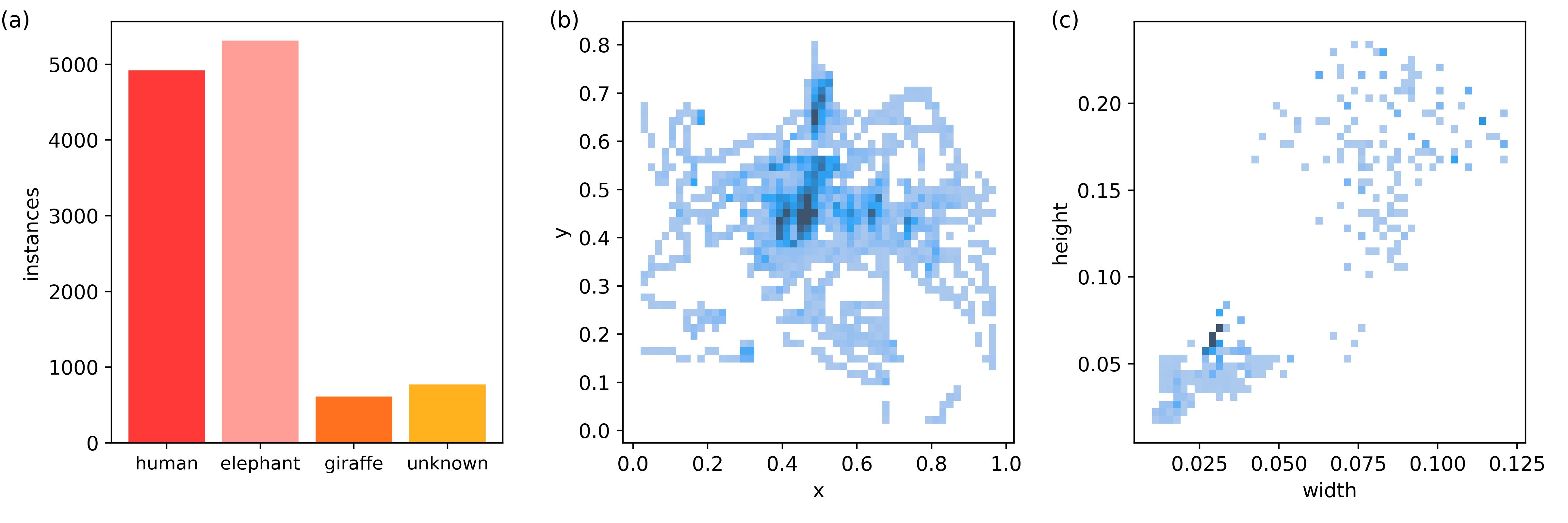}
\caption{Label distribution of the ground-truth boxes on the test set.\label{combined_plots2.jpg}}
\end{figure}

Fig.~\ref{combined_plots.jpg} shows the distribution of labels in the train set. Fig.~\ref{combined_plots.jpg}(a) is a histogram showing the distribution of instances in each category. The x-axis in the figure represents the different categories in the dataset, and the y-axis represents the number of instances. Fig.~\ref{combined_plots.jpg}(b) is a scatter plot showing the distribution density of the position of the true box in the entire image, where "x" represents the ratio of the target's horizontal center point to the image width, and "y" represents the ratio of the target's vertical center point to the image height. The dark blue areas show high data point density, while light blue areas indicate low density with sparse data distribution. It can be seen that the data points are mainly distributed in the center of the picture. Fig.~\ref{combined_plots.jpg}(c) shows the distribution density of the width and height of the true box in the entire image, where "width" signifies the ratio of the target's width to the image width, and "height" signifies the ratio of the target's height to the image height. The darker the color, the more concentrated the objects within the width and height range. Its data distribution reveals the richness of small targets, with the width and height concentrated at approximately 0.05 and 0.15, which is consistent with the narrow and tall characteristics of elephants that exist in large numbers in the dataset. Fig.~\ref{combined_plots2.jpg} shows the label distribution in the test set, and its distribution pattern is consistent with the above discussion.

\subsubsection{Evaluation Criteria}

We employed a range of evaluation metrics to assess model performance, including precision (P), recall (R), average precision (AP), mean average precision (mAP), model parameter count, and frame rate (FPS). Precision evaluates the accuracy of positive predictions, while recall assesses the model's ability to identify all relevant instances. The F1 score is the harmonic mean of precision and recall. The AP value evaluates the prediction accuracy of a single category, while mAP evaluates the average precision across all categories, serving as a crucial metric for evaluating the overall performance of a detection model. The model's size is determined by the number of parameters, which signifies the complexity of the model. FPS quantifies the processing speed of the model, representing the number of frames processed per second. It should be emphasized that if not otherwise stated, precision and recall are measured at an IOU threshold of 0.5 and a confidence threshold corresponding to the maximum F1 score.

These indicators together constitute a comprehensive evaluation system for model performance. The calculation process is as follows:
\begin{equation}
\ { P }=\frac{T P}{T P+F P}
\end{equation}
\begin{equation}
\ { R }=\frac{T P}{T P+F N}
\end{equation}
\begin{equation}
A P=\int_{0}^{1} P(R) d R
\end{equation}
\begin{equation}
\ { F1}=\frac{2 * {P} * {R}}{{P}+{R}}=\frac{2 {TP} ^ {2}}{2 {TP}+{FP}+{FN}}
\end{equation}
\begin{equation}
m A P=\frac{1}{n} \sum_{i=1}^{n} A P_{i}
\end{equation}
True Positive (TP), False Positive (FP), and False Negative (FN) denote correctly identified, incorrectly identified, and missed samples, respectively. N represents the number of categories to be classified.

\subsubsection{Implementation Details}
All experiments reached convergence after the model was trained for 200 epochs. The model utilized an input image size of 640x640 and employed the SGD optimizer with specific parameters: a batch size of 120, momentum of 0.937, and weight decay of 0.0005. To stabilize training, we initiated a 3-epoch warm-up phase with an optimizer momentum of 0.8. Following the warm-up training, the learning rate was decayed using a cosine annealing function, with initial and minimum rates set at 0.001 and 0.00001, respectively. The experimental setup consists of an Intel i9-10900K CPU, an NVIDIA GeForce RTX 3090 GPU with 24GB memory, and the Ubuntu 20.04 operating system. The frameworks utilized are Python 3.11.0, PyTorch 2.2.2, and CUDA 12.0.

\begin{table*}[!t]
\centering
  \caption{Ablation experiments based on the BIRDSAI TIR UAV  dataset.\label{ablation_table}}
    \begin{tabular}{ccccccccccccc}
      \hline
      \textbf{Model} & \textbf{ALSS} & \textbf{Focus} & \textbf{Pool-Conv} & \textbf{LCA} & \textbf{FineSIOU} & \textbf{CA} & \textbf{P} & \textbf{R} & \textbf{mAP0.50} & \textbf{Params(m)} & \textbf{FPS} \\
      \hline
YOLOv8-N'   &     &      &      &      &      &      & 0.869     & 0.834    & 0.874      & 1.795      & 215.4        \\
M1          &\checkmark  &   &  &      &      &      & 0.889     & 0.840    & 0.877      & 1.482      & 227.8        \\
M2          & \checkmark & \checkmark  &  &   &  &   & 0.882     & 0.848    & 0.887      & 1.483      & 221.9        \\
M3          & \checkmark & \checkmark  & \checkmark  & & & & 0.886 & 0.834    & 0.886      & 1.433      & 228.7        \\
M4          & \checkmark & \checkmark  & \checkmark  & \checkmark &  & & 0.883  & 0.854 & 0.889 & 1.452 & 225.6        \\
M5(ALSS-YOLO)  & \checkmark & \checkmark & \checkmark & \checkmark & \checkmark & & 0.887  & 0.857 & 0.891 & 1.452 & 223.7        \\
M6             & \checkmark & \checkmark & \checkmark &  &\checkmark  & \checkmark &0.891  & 0.847 & 0.888 & 1.453 & 224.2        \\
\hline
\end{tabular} \\
\end{table*}

\begin{table*}[!t]
\centering
\caption{The AP0.50 values of various categories in the ablation experiment based on the BIRDSAI TIR UAV dataset.\label{ablation_table2}}
\begin{tabular}{ccccccccccccccc}
\hline
\textbf{Model} & \textbf{human} & \textbf{elephant} & \textbf{giraffe} & \textbf{unknown} & \textbf{mAP0.50} \\
\hline
YOLOv8-N'      & 0.905          & 0.975             & 0.780            & 0.837            & 0.874                \\
M1             & 0.902          & 0.979             & 0.779            & 0.850            & 0.877                \\
M2             & 0.910          & 0.980             & 0.813            & 0.836            & 0.887                \\
M3             & 0.917          & 0.979             & 0.794            & 0.863            & 0.886                \\
M4             & 0.918          & 0.982             & 0.796            & 0.860            & 0.889                \\
M5(ALSS-YOLO)  & 0.923          & 0.978             & 0.806            & 0.855            & 0.891                \\
M6             & 0.900          & 0.976             & 0.798            & 0.878            & 0.888                \\
\hline
\end{tabular}
\end{table*}

\begin{figure*}[!t]
\centering
\includegraphics[width=13cm]{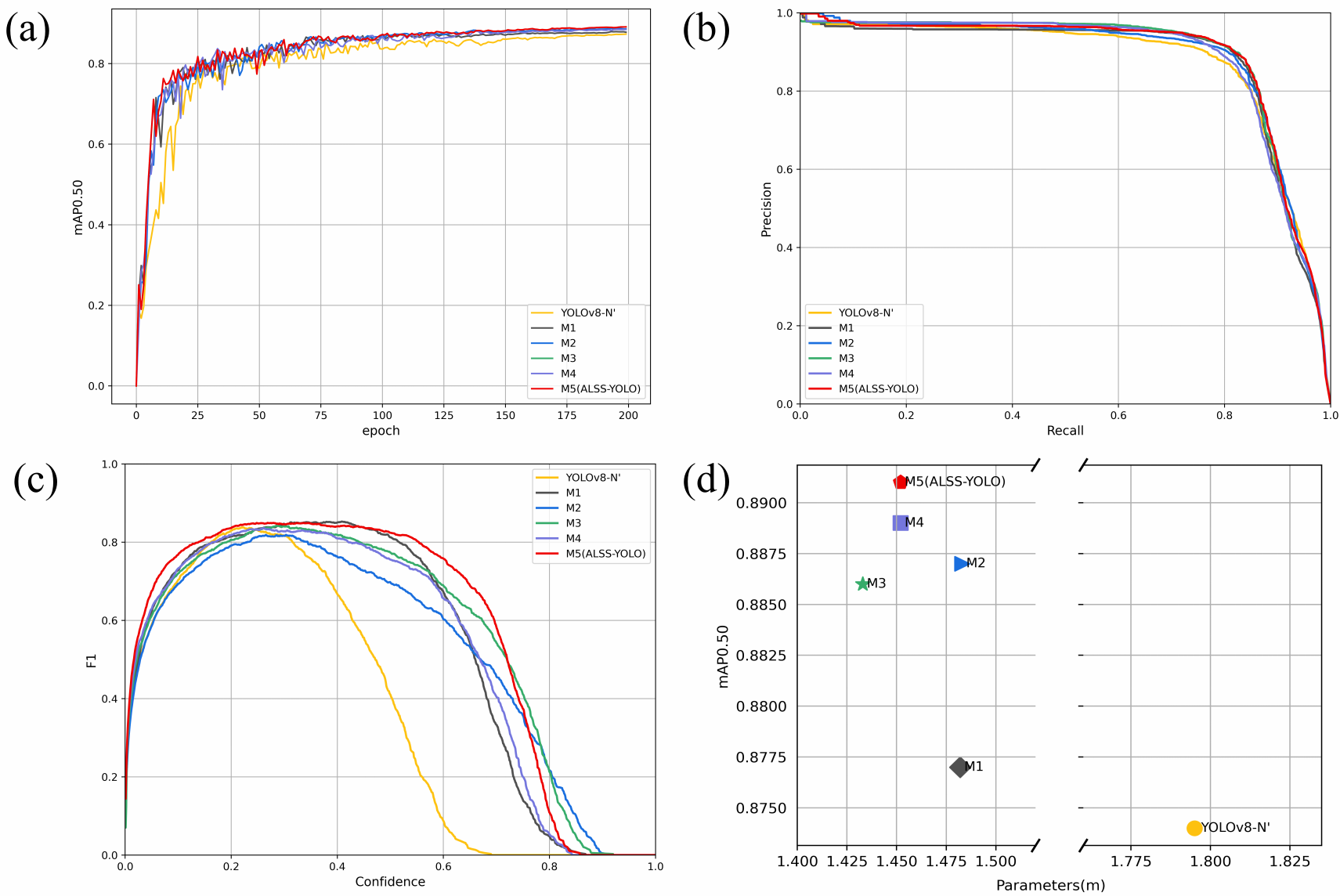}
\caption{Performance comparison chart of ablation experiment based on BIRDSAI TIR UAV dataset: (\textbf{a}) mAP0.50 curve changing with training rounds; (\textbf{b}) accuracy and recall rate fit curve; (\textbf{c}) curve of F1 value changing with confidence threshold; (\textbf{d}) scatter plot of the relationship between mAP0.50 and parameter quantity.\label{ablation_curve}}
\end{figure*}

\begin{figure*}[!t]
\centering
\includegraphics[width=15cm]{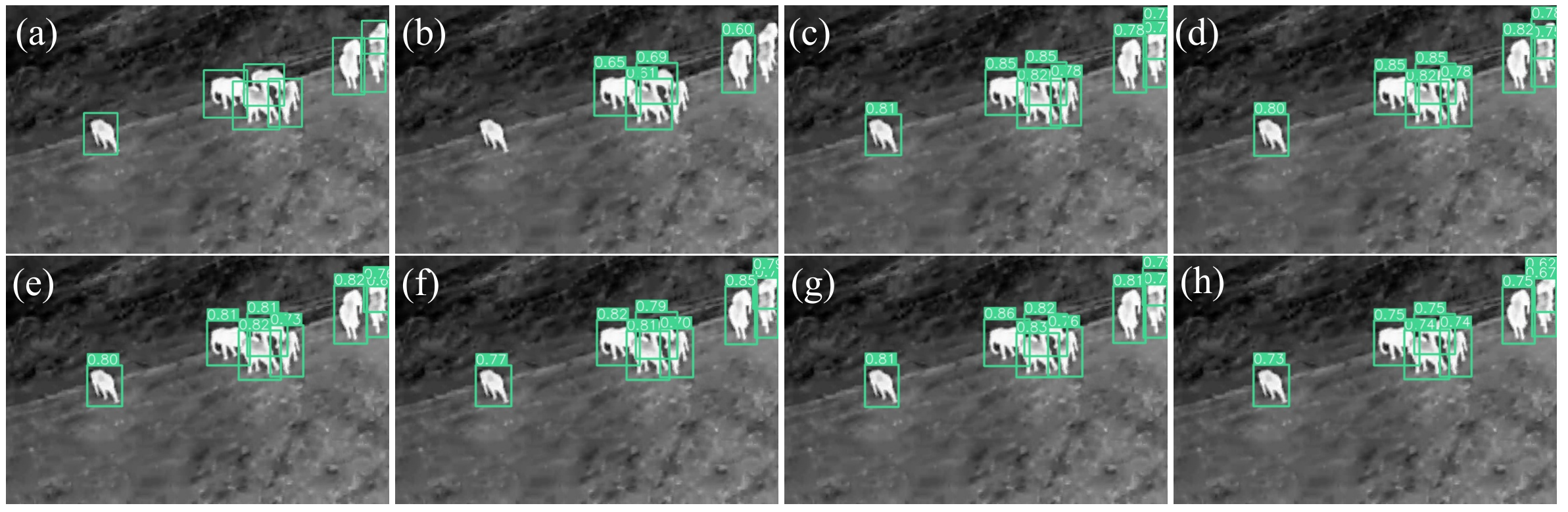}
\caption{Detection examples of different ablation experiments in the BIRDSAI TIR UAV dataset: (\textbf{a) ground truth; (\textbf{b}) YOLOv8-N'; (\textbf{c}) M1; (\textbf{d}) M2; (\textbf{e}) M3; (\textbf{f}) M4; (\textbf{g}) M5(ALSS-YOLO) (\textbf{h}) M6.} \label{ablation_virsion}}
\end{figure*}

\begin{figure*}[!t]
\centering
\includegraphics[width=15cm]{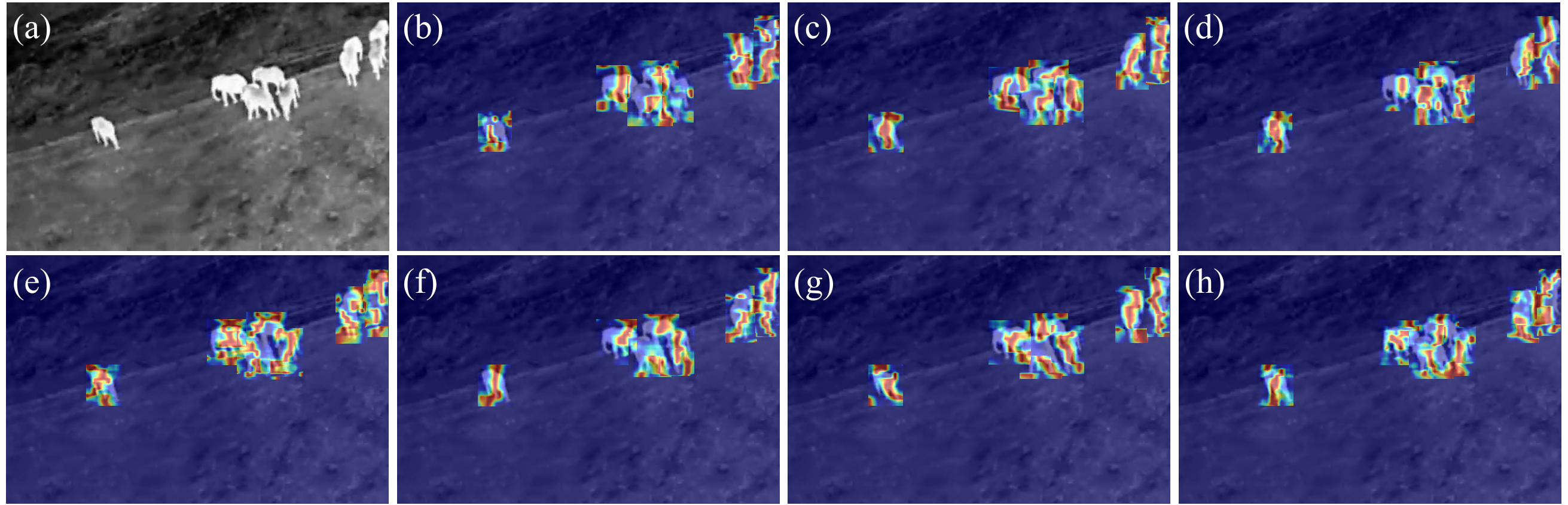}
\caption{Example feature activation maps of detections from different ablation experiments in the BIRDSAI TIR UAV dataset: (\textbf{a) ground truth; (\textbf{b}) YOLOv8-N'; (\textbf{c}) M1; (\textbf{d}) M2; (\textbf{e}) M3; (\textbf{f}) M4; (\textbf{g}) M5(ALSS-YOLO) (\textbf{h}) M6.} \label{ablation_virsion2}}
\end{figure*}

\subsection{Ablation Experiments}

To assess the effects of the enhanced strategies proposed in this research, a series of ablation experiments were conducted to quantify the outcomes. as depicted in Table~\ref{ablation_table}, Table~\ref{ablation_table2} and Fig.~\ref{ablation_curve}, indicate that each technical enhancement led to performance gains. Building on the YOLOv8-n baseline model, to maintain the approximate consistency of the model parameter quantity, we adjusted the width hyperparameter of the model from 0.25 to 0.18 and marked it as YOLOv8-N' (although we also investigated the adjustment of the depth hyperparameter, the results were not good, with a mAP0.50 of 0.863). This study proposes a network architecture centered on the ALSS module, incorporating a single-channel focus module. The original model’s $3 \times 3$ convolutions with a stride of 2 in layers 16 to 19 and 20 to 23 are replaced with max pooling and $1 \times 1$ pointwise convolutions. Furthermore, an LCA module is introduced, and the loss function is modified accordingly. 
Additionally, a control experiment comparing the performance of CA and LCA modules is conducted.

To investigate the effectiveness of the ALSS module as the core component of the model, this study modified YOLOv8-N' by replacing the original C2F module with the ALSS module and adjusting the channel ratio of various feature scales directed to the detection head. The new network structure is named M1. According to the data in Table~\Ref{ablation_table} (rows 1 and 2), compared to YOLOv8-N', the M1 model achieved a 0.3 percentage point increase in the mAP0.50 metric while reducing the number of parameters by 17.4\%. It is particularly noteworthy that the precision significantly increased from 0.869 to 0.889. Fig.~\Ref{ablation_curve}(a) reveals that the M1 model has a faster convergence speed compared to YOLOv8-N', and Fig.~\ref{ablation_curve}(c) shows that the F1 score of YOLOv8-N' significantly decreases under high confidence threshold settings, a phenomenon further confirmed in subsequent Fig.~\ref{ablation_virsion}. These results fully demonstrate the lightweight and efficient nature of the ALSS module.

Furthermore, to confirm the efficacy of the single-channel focus module in augmenting the model's capability to discern intricate patterns and structures, this study introduced the module at the first layer of the M1 model, resulting in the M2 network structure. Table~\ref{ablation_table} (rows 2 and 3) shows that with roughly the same amount of parameters, the M2 model’s mAP0.50 was improved by 1\%, at the same time, Table~\ref{ablation_table2} (rows 2 and 3) points out that except for the AP value of the unknown category which decreased slightly, the AP values ​​of other categories increased.

To further diminish the number of parameters and the model's complexity, we replace the $3 \times 3 $ convolutional layers (stride 2) of layers 16 to 19 and 20 to 23 of the M2 model with max-p cooling layers of stride 2 and pointwise convolution structure, denoted as M3. Downsampling is performed through max-pooling improving the spatial invariance of the model. Compared with the convolution layer with a convolution kernel size of 3, 
pointwise convolution, with fewer parameters and computational costs, facilitates altering the depth of the input feature map to achieve cross-channel information integration. According to the data in Table~\ref{ablation_table} and Table~\ref{ablation_table2} (rows 3 and 4), the number of parameters of the M3 model is reduced by 3.4\% compared with M2. At the same time, when mAP0.50 only decreases by 0.1\%, and the AP values for each category are similar. Fig.~\Ref{ablation_curve} also shows that the M3 model maintains approximate performance under multiple evaluation indicators.

Additionally, to assess the effect of the LCA module in enhancing feature discrimination capability and the overall performance of the network while maintaining relatively low complexity and high computational efficiency, this study integrated the LCA module at the 15th layer of the M3 model, thus developing the M4 model. From the comparison between Table~\ref{ablation_table} and Table~\ref{ablation_table2} (rows 4 and 5), Although the number of model parameters increased by 1.3\%, the mAP increased by 0.3\%. Considering that the baseline mAP value had reached 88.6\%, this increase is quite significant. At the same time, the recall also improved from 0.834 to 0.854, with AP values for various categories having their respective gains and losses.

Modifying the loss function generally influences the training phase exclusively and does not have an impact on the inference time of the network. This paper presents the FineSIOU loss function as a method to improve the accuracy of the detection bounding boxes. The data from Table~\ref{ablation_table} and ~\ref{ablation_table2} (rows 5 and 6) as well as the trends in Fig.~\Ref{ablation_curve} show that the introduction of the new loss function brought a 0.2\% gain to the mAP0.50, and it also performed best in terms of model convergence speed, PR curve, and F1 score. The model achieved an optimized balance between accuracy and speed, realizing higher accuracy on UAV platforms with limited computing resources. The separation of the angular cost as an independent term in the FineSIOU loss function accelerated convergence, and given the large number of small targets present in the BIRDSAI database, the model’s capability for small object detection has also been empirically proven.

Besides, we conducted comparative experiments with the CA module, as shown in Tables~\ref{ablation_table} and~\ref{ablation_table2} (Rows 6 and 7). The experiments demonstrated that the LCA module achieved superior results on the BIRDSAI dataset, with the M5 model outperforming M6 by 0.3\% mAP50. We also observed this phenomenon on the ISOD\cite{CE-RetinaNet}dataset. These findings suggest the LCA module's potential for enhanced performance in certain scenarios. Future work will extend these evaluations to additional datasets to further validate the performance differences and explore the applicability of each module across various contexts.

The ablation experiment shown in Fig.~\ref{ablation_virsion} was performed based on detection results with a confidence level greater than 0.6. As can be seen from Fig.~\ref {ablation_curve}(c), when the confidence level is 0.6, the F1 value is only about 0.1, which shows that the recall rate of the model is significantly reduced. As illustrated in Fig.~\ref{ablation_virsion}(a), there are numerous missed detections, but in our improved subsequent model, the missed detections have been eliminated. In addition, the ALSS-YOLO detection frame has the highest overall confidence.

Fig.~\ref{ablation_virsion2} provides a visualization of the feature activation maps in the detection area after the 10th, 12th, 14th, and 16th layer modules. These visualizations reveal that the network, especially after integrating the LCA module, effectively focuses on the entire object being detected. Even in the relatively shallow layers, the network can concentrate on the semantic information of the target, demonstrating excellent feature abstraction. This highlights the LCA module's ability to improve feature representation and discrimination, contributing to the overall effectiveness of the detection process.

\begin{table*}[!t]
\centering
\caption{Comparative experiments were conducted after adjusting the width and depth hyperparameters to achieve similar parameter sizes.\label{params-comparison}}
\begin{tabular}{ccccccccccccc}
\hline
\textbf{Model} & \textbf{P} & \textbf{R} & \textbf{mAP0.50} & \textbf{human} & \textbf{elephant} & \textbf{giraffe} & \textbf{unknown} & \textbf{Params(m)} & \textbf{FPS}              \\
\hline
YOLOv8-n       & 0.885          & 0.858           & 0.894                & 0.921          & 0.982              & 0.816            & 0.859            & 3.006               & 213.1        \\
ALSS-YOLO-m    & 0.897          & 0.863           & 0.903                & 0.931          & 0.983             & 0.803            & 0.882            & 2.924              & 215.5        \\
YOLOv8-n''     & 0.879          & 0.849           & 0.884                & 0.922          & 0.980             & 0.798            & 0.837            & 2.364               & 207.8        \\
ALSS-YOLO-s    & 0.898          & 0.856           & 0.895                & 0.928          & 0.983             & 0.787            & 0.861            & 2.226               & 212.3\\       
\hline
\end{tabular}
\end{table*}

\begin{table*}[!t]
\centering
  \caption{Comparative experiments on different network structure selection strategies.\label{structure-comparison}}
    \begin{tabular}{ccccccccccccc}
      \hline
      \textbf{Model} & \textbf{P} & \textbf{R} & \textbf{mAP0.50} & \textbf{human} & \textbf{elephant} & \textbf{giraffe} & \textbf{unknown} & \textbf{Params(m)} & \textbf{FPS}              \\
      \hline
      ALSS-YOLO      & 0.887          & 0.857           & 0.891                & 0.923          & 0.978             & 0.806            & 0.855            & 1.452              & 223.7        \\
      ALSS-YOLO'     & 0.886          & 0.840           & 0.887                & 0.913          & 0.981             & 0.801            & 0.853            & 1.460              & 218.9        \\
      ALSS-YOLO"    & 0.897          & 0.818           & 0.876                & 0.923          & 0.978             & 0.806            & 0.855            & 1.453              & 226.6       \\  
      \hline
    \end{tabular}
\end{table*}

\begin{table*}[!t]
\centering
\caption{Comparative Performance Metrics of ALSS-YOLO with LCA Module Applied at Different Layers.\label{LCA-comparison}}
\begin{tabular}{ccccccccccccc}
\hline
\textbf{Model} & \textbf{P} & \textbf{R} & \textbf{mAP0.50} & \textbf{human} & \textbf{elephant} & \textbf{giraffe} & \textbf{unknown} & \textbf{Params(m)} & \textbf{FPS}              \\
\hline
ALSS-YOLO(LCA@L8)       & 0.889          & 0.852           & 0.887                & 0.916          & 0.977             & 0.783            & 0.871            & 1,426               & 231.8        \\
ALSS-YOLO(LCA@L18)      & 0.887          & 0.845           & 0.889                & 0.909          & 0.977             & 0.796            & 0.874            & 1,415              & 222.1        \\
ALSS-YOLO(LCA@L22)      & 0.883          & 0.837           & 0.875                & 0.912          & 0.978             & 0.758            & 0.852            & 1,449               & 232.8        \\
ALSS-YOLO               & 0.887          & 0.857           & 0.891                & 0.923          & 0.978             & 0.806            & 0.855            & 1.452               & 223.7       \\
\hline
\end{tabular}
\end{table*}

\begin{table*}[!t]
\centering
  \caption{Comparative experiments based on the BIRDSAI TIR UAV dataset.\label{comparison-table}}
    \begin{tabular}{ccccccccccccc}
      \hline
      \textbf{Model} & \textbf{P} & \textbf{R} & \textbf{mAP0.50} & \textbf{human} & \textbf{elephant} & \textbf{giraffe} & \textbf{unknown} & \textbf{Params(m)} & \textbf{FPS}              \\
      \hline
      YOLOv3-tiny\cite{YOLOv3}     & 0.827          & 0.732           & 0.783                & 0.835          & 0.968             & 0.578            & 0.751            & 1.534               & 304.6        \\
      YOLOX-nano\cite{Yolox}       & 0.849          & 0.837           & 0.865                & 0.831          & 0.964             & 0.805            & 0.862            & 0.912              & 201.2        \\
      YOLOv5-n'       & 0.869          & 0.837           & 0.876                & 0.909          & 0.978             & 0.805            & 0.817            & 1.518               & 207.3        \\
      YOLOv6-n'\cite{YOLOv6}       & 0.850          & 0.830           & 0.864                & 0.896          & 0.976             & 0.729            & 0.854            & 1.793               & 336.7        \\
      YOLOv8-ghost    & 0.877          & 0.847           & 0.877                & 0.910           & 0.979             & 0.782            & 0.835            & 1.712               & 163.5        \\
      YOLOv8-AM\cite{YOLOv8-AM}       & 0.875         & 0.848           & 0.886                & 0.924          & 0.978             & 0.807            & 0.835            & 1.583               & 220.9        \\
      YOLOv8-p2       & 0.885          & 0.872           & 0.894                & 0.939          & 0.981             & 0.789            & 0.866            & 2.927               & 170.4        \\
      MASK-RCNN-r18\cite{Mask}  & 0.635*          & 0.697*           & 0.901                & 0.885          & 0.973             & 0.824            & 0.906            & 30.919             & 38.8          \\
      FASR-RCNN-r18\cite{Faster} & 0.640*          & 0.698*           & 0.907                & 0.892          & 0.974             & 0.839            & 0.922            & 28.684             & 34.8         \\
      RTMDet-tiny\cite{RTMDet}     & 0.626*          & 0.670*           & 0.887                & 0.907          & 0.973             & 0.781            & 0.886            & 4.874              & 218.2        \\
      VarifocalNet\cite{VarifocalNet}    & 0.592*          & 0.674*          & 0.869                & 0.908          & 0.977             & 0.815            & 0.777            & 32.716             & 14.5         \\
      CE-RetinaNet       & 0.642*          & 0.694*           & 0.918                & 0.895          & 0.966             & 0.753            & 0.841            & 55.823              & 6.1    \\
      ALSS-YOLO       & 0.887          & 0.857           & 0.891                & 0.923          & 0.978             & 0.806            & 0.855            & 1.452              & 223.7    \\
      \hline
    \end{tabular}\\
  \footnotesize{The symbol ' indicates that the width hyperparameter of the model is adjusted to approximate the number of parameters in the ALSS-YOLO model. * indicates that these algorithms were reproduced following the configurations in \cite{mmdetection}, with the default settings of confidence threshold = 0.3 and IoU threshold = 0.5.}
\end{table*}

\begin{figure*}[!t]
\centering
\includegraphics[width=13cm]{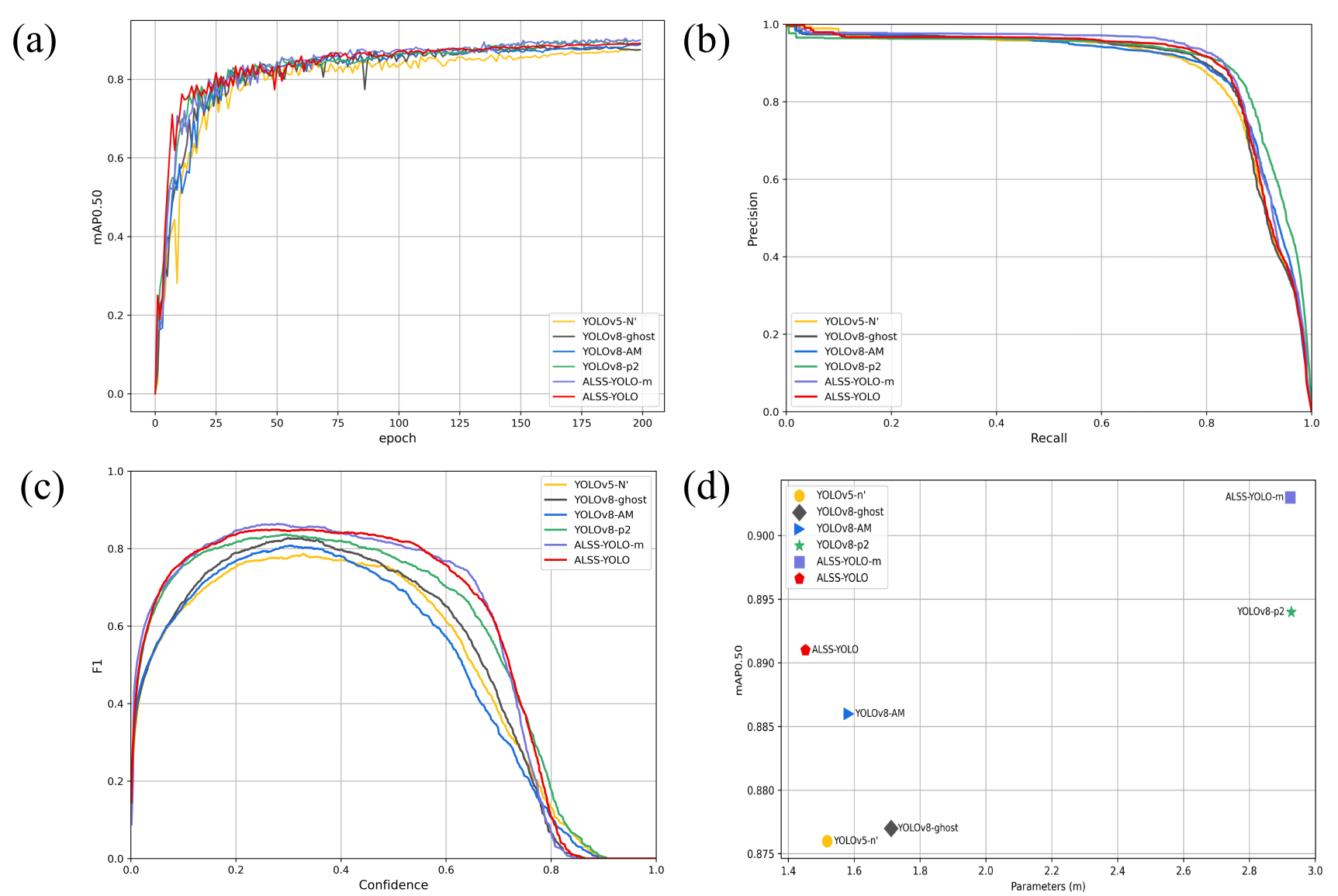}
\caption{Performance comparison chart of comparison experiment based on BIRDSAI TIR UAV dataset: (\textbf{a}) mAP0.50 curve changing with training rounds; (\textbf{b}) fitting curve of precision and recall rate; (\textbf{c}) curve of F1 value changing with confidence threshold; (\textbf{d}) scatter plot of the relationship between mAP0.50 and parameter quantity.\label{comparison_curve}}
\end{figure*}

\begin{figure*}[!t]
\centering
\includegraphics[width=13cm]{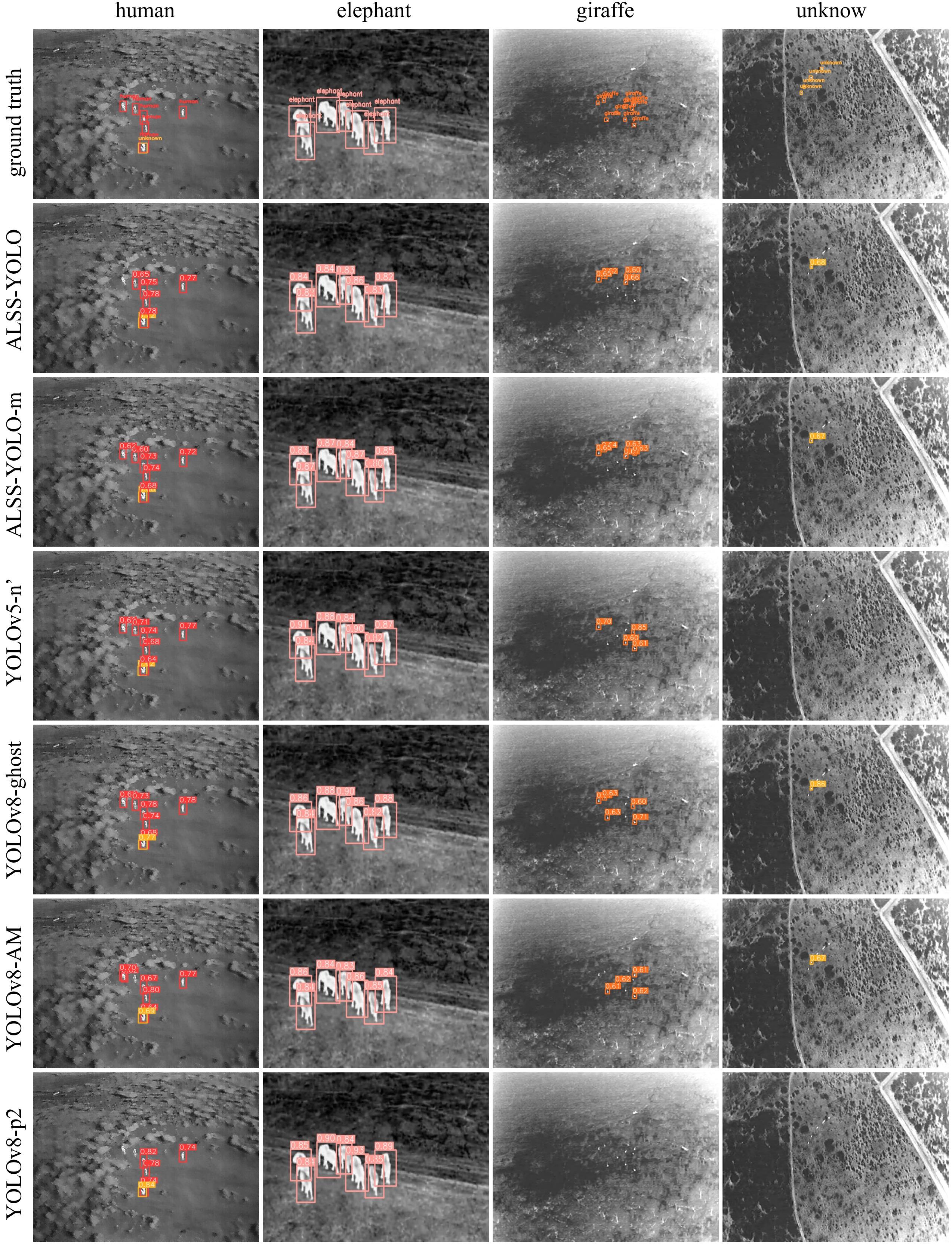}
\caption{Detection results of examples from different categories in the test set.\label{comparison_virsion.png}}
\end{figure*}

\begin{figure*}[!t]
\centering
\includegraphics[width=13cm]{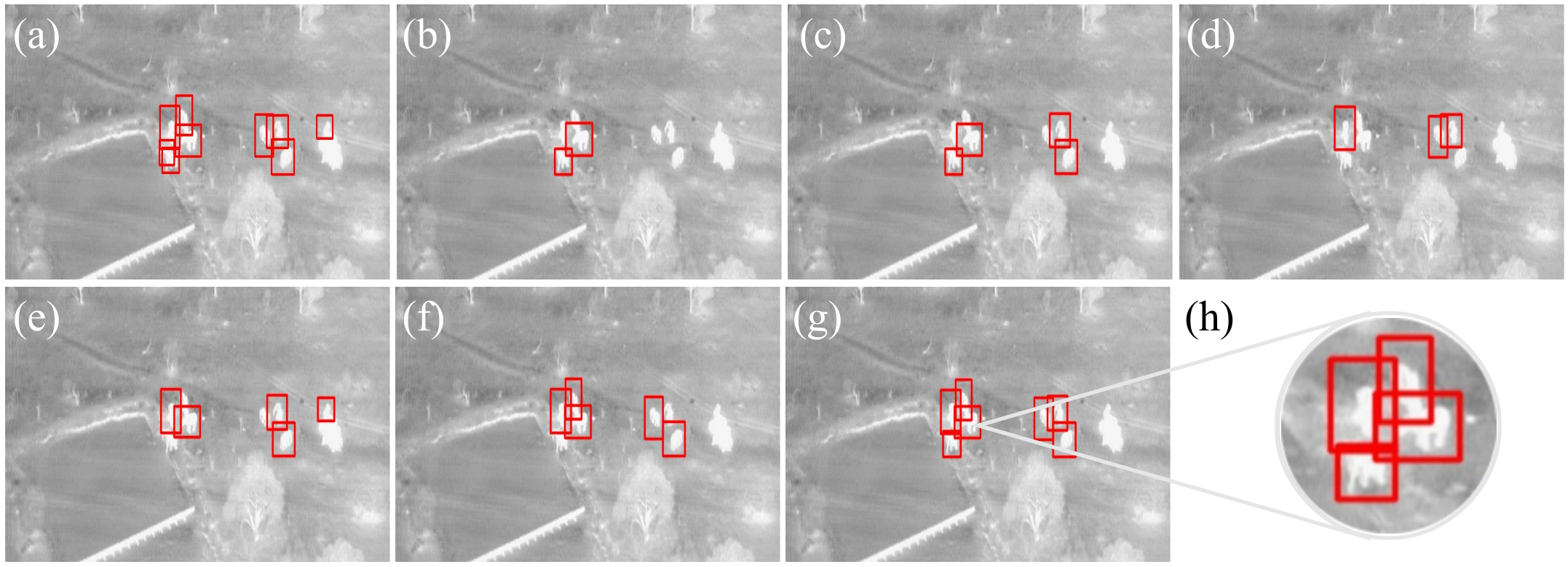}
\caption{Detection examples of different algorithms in high-noise scenarios: (\textbf{a) ground truth; (\textbf{b}) YOLOv5-n'; (\textbf{c}) YOLOv8-ghost; (\textbf{d}) YOLOv8-AM; (\textbf{e}) YOLOv8-p2;(\textbf{f}) ALSS-YOLO-n; (\textbf{g}) ALSS-YOLO-m; (\textbf{h}) Detection Details of ALSS-YOLO-m.}\label{comparison_noise}}
\end{figure*}

\begin{figure*}[!t]
\centering
\includegraphics[width=13cm]{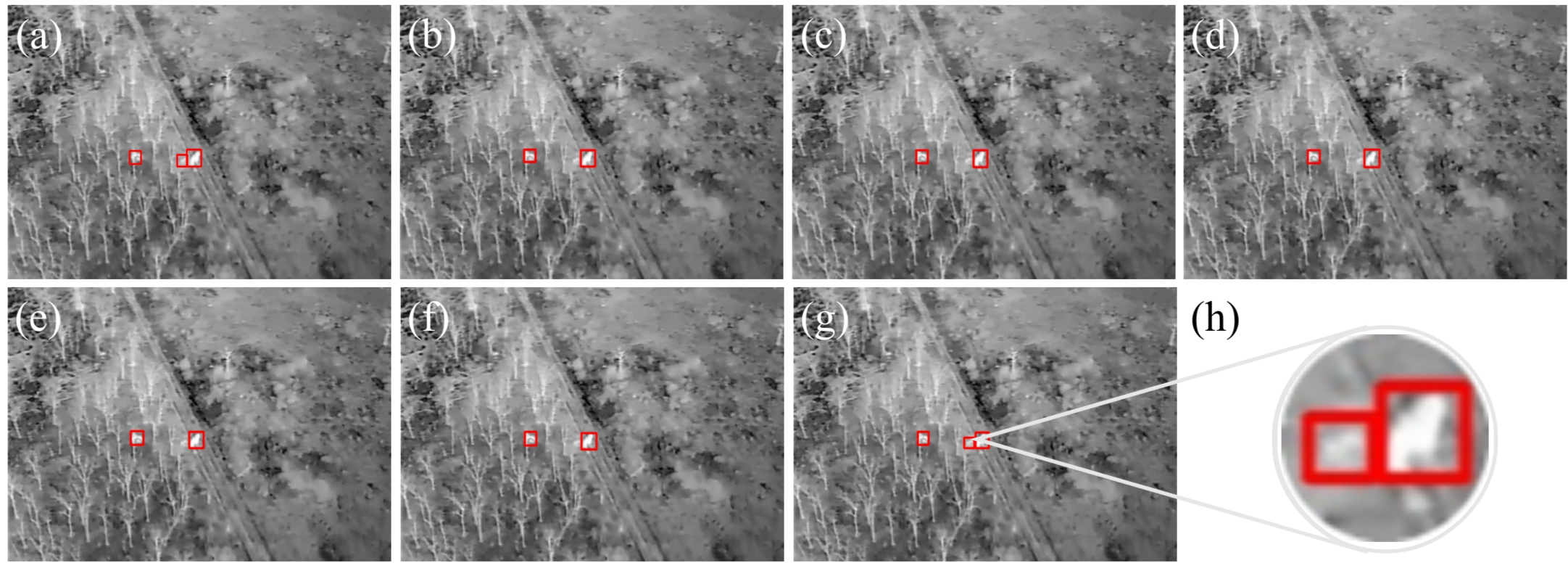}
\caption{Detection examples of different algorithms in a heavily occluded scene: (\textbf{a) ground truth; (\textbf{b}) YOLOv5-n'; (\textbf{c}) YOLOv8-ghost; (\textbf{d}) YOLOv8-AM; (\textbf{e}) YOLOv8-p2;(\textbf{f}) ALSS-YOLO-n; (\textbf{g}) ALSS-YOLO-m; (\textbf{h}) Detection Details of ALSS-YOLO-m.}\label{comparison_occlusion}}
\end{figure*}

\subsection{Comparison Experiments}

To substantiate the superior performance of our model, we meticulously adjusted the width and depth hyperparameters of the ALSS-YOLO framework, resulting in two novel variants, designated ALSS-YOLO-s, and ALSS-YOLO-m. Concurrently, the width hyperparameter of the YOLOv8-n model was fine-tuned to generate YOLOv8-n''. As shown in Table~\ref{params-comparison}, our ALSS-YOLO-m model achieved a mAP0.50 gain of 0.9\% over the YOLOv8-n, while the ALSS-YOLO-s attained a 1.1\% mAP increase compared to YOLOv8-n'', despite similar parameter counts. It is noteworthy that although our model’s AP for detecting the giraffe category was slightly inferior to that of YOLOv8, it outperformed in error detection across all other categories.

To validate the strategy of setting the parameter $\alpha$ as discussed in Section 3.1 and the network selection strategy illustrated in parts A of Fig.~\ref{ALSS-S=1} and ~\ref{ALSS-S=2} in the ALSS module, we conducted experiments as presented in Table~\ref{structure-comparison}.
Specifically, ALSS-YOLO’ implements a contrary $\alpha$ value strategy to that of ALSS-YOLO. Higher $\alpha$ values were assigned to lower-level feature layers to ensure the majority of feature channels pass through the network structure shown in part A of Fig.~\ref{ALSS-S=1} and ~\ref{ALSS-S=2}. Conversely, lower $\alpha$ values were set for higher-level feature layers. ALSS-YOLO’’ maintained the same $\alpha$ values as ALSS-YOLO but adopted an opposite strategy for network structure selection. Namely, it utilized identity connections for feature extraction at lower-level feature layers and convolutional operations at higher-level feature layers. The data from Table~\ref{structure-comparison} indicate that the mAP0.50 of ALSS-YOLO is 0.4\% higher than that of ALSS-YOLO’ and 1.5\% higher than that of ALSS-YOLO''. These findings validate the discussion in Section 3.1 and demonstrate that the network structure selection strategy of ALSS-YOLO can more effectively learn and represent the features of the input data, thereby enhancing model performance.

To investigate the impact of integrating the LCA module at different layers, we conducted a series of comparative experiments using the ALSS-YOLO model. Specifically, we analyzed the performance of the model when the LCA module was placed at the 8th, 18th, and 22nd layers, respectively. These configurations are denoted as ALSS-YOLO(LCA@L8), ALSS-YOLO(LCA@L18), and ALSS-YOLO(LCA@L22). The original ALSS-YOLO model is also included for comparison. The results of these experiments are summarized in Table~\ref{LCA-comparison}. 
The findings reveal that ALSS-YOLO achieved the highest mAP@0.50 and demonstrated strong competitiveness across other metrics. Conversely, ALSS-YOLO(LCA@L22) recorded the lowest mAP@0.50, indicating a negative impact on overall model performance. These results further validate the rationale for the LCA module placement as discussed in Section 3.2.

To validate the performance of the ALSS-YOLO model, we conducted a comparison with other state-of-the-art object detection models on the BIRDSAI TIR UAV dataset. These models include YOLOv3-tiny, YOLOX-nano, YOLOv5-n', YOLOv6-n', YOLOv8-ghost, YOLOv8-AM, YOLOv8-p2, MASK-RCNN-r18, FASR-RCNN-r18, RTMDet-tiny, VarifocalNet and CE-RetinaNet. To ensure a fair comparison of each model's performance, all models are trained in the same training environment. The comparison results of ALSS-YOLO with other models are presented in Table~\ref{comparison-table}, while Fig.~\ref{comparison_curve} shows the performance comparison of ALSS-YOLO and the best models.

In the comparative experimental evaluation on the BIRDSAI TIR UAV dataset, the proposed ALSS-YOLO model achieved a remarkable mAP0.50 of 89.1\% with only 1.452 million parameters, demonstrating its efficiency for UAV-based applications. While the CE-RetinaNet, designed specifically for infrared wildlife detection, attained a higher mAP0.50 of 91.8\%, its large parameter count of 55.823 million makes it unsuitable for deployment on UAVs with limited computational resources.
In contrast, YOLOX-nano, despite having the smallest parameter count (0.912 million), showed a 2.6\% lower detection precision compared to ALSS-YOLO. Our detailed analysis, as depicted in Fig.\ref{comparison_curve}, highlights that while YOLOv8-p2 slightly outperforms our model, ALSS-YOLO-m, with a comparable parameter count, surpasses YOLOv8-p2 by 0.9\%, achieving a mAP0.50 of 90.3\%. The superior recall rates of ALSS-YOLO, illustrated in Fig.\ref{comparison_curve}(b,c), further underscore its robustness.

These findings, clearly demonstrate ALSS-YOLO's effectiveness in handling challenges such as image blurring and target overlap in TIR wildlife detection tasks. Our model's balance between accuracy and computational efficiency makes it highly suitable for UAV-based monitoring.

To showcase the versatility of our model across different categories, we present the detection results of various species under typical scene conditions. The test outcomes are illustrated in Fig.~\ref{comparison_virsion.png}. ALSS-YOLO-m displayed superior overall detection performance, with no missed detections or false alarms. While ALSS-YOLO had a minor missed detection in one scenario, it still outperformed other models like YOLOv8-p2, which suffered significant missed detections in certain challenging scenarios, likely due to specific scene conditions or image blurriness, as detailed in Table\ref{comparison-table}. These analyses confirm that ALSS-YOLO offers a robust solution for TIR wildlife detection in UAV applications.

Fig.\ref{comparison_noise} and \ref{comparison_occlusion} illustrate the model's detection performance under highly challenging conditions. In Fig.\ref{comparison_noise}, depicting a scene with significant noise interference, the ALSS-YOLO-m model achieved the highest recall rate, demonstrating superior detection performance. Meanwhile, the ALSS-YOLO-n model also delivered highly competitive results, underscoring its effectiveness in challenging conditions. Fig.\ref{comparison_occlusion}, which presents a scenario with severe occlusion, reveals that only the ALSS-YOLO-m model was able to detect all the targets, highlighting its robustness in such difficult situations.

\subsection{Experimental Evaluation of the ISOD dataset}

In the experimental evaluation of the ISOD\cite{CE-RetinaNet} dataset, we validated the generalization capabilities of the ALSS-YOLO model. The results, as shown in Table~\ref{comparison-table2}, demonstrate that ALSS-YOLO outperforms several well-known object detectors in terms of AP50 while maintaining a significantly lower parameter count and faster inference speed. The results for the ALSS-YOLO experiment were obtained from our study, while the data for other models in this comparison were sourced from the CE-RetinaNet.

Specifically, ALSS-YOLO-m achieved an impressive AP50 of 80.3\%, surpassing many models such as RetinaNet (ResNet-101)\cite{RetinaNet} and CE-RetinaNet (ResNeXt-101)\cite{CE-RetinaNet} while requiring only a fraction of their parameters. Additionally, ALSS-YOLO-m maintained a high recall rate, further confirming its effectiveness for real-time applications in UAV-based monitoring on the ISOD dataset. These findings underscore ALSS-YOLO's balance between accuracy and efficiency, making it a robust solution for scenarios that demand both high precision and real-time processing.

\begin{table}[!h]
\caption{Comparative experiments based on the ISOD TIR UAV  dataset.\label{comparison-table2}}
\begin{tabular}{p{2cm}|p{2cm}|p{0.5cm}|p{0.5cm}|p{0.5cm}|p{0.5cm}}
\hline 
\textbf{Model}& \textbf{Backbone}&\textbf{AP50}&\textbf{R}&\textbf{P(m)} & \textbf{FPS}\\
\hline 
Cascade RCNN\cite{Cascade_R_CNN}	& ResNet-101	& 0.548		& 0.617		& 87.92	& 5.8	\\
Dynamic RCNN\cite{Dynamic_R_CNN}	& ResNet-101	& 0.524	  & 0.617		& 60.11	& 60	\\
GHM\cite{GHM}	          & ResNet-101 	& 0.409		& 0.553		& 55.1	& 6.1	\\
ATSS\cite{ATSS}	        & ResNet-101 	& 0.438		& 0.637		& 50.88	& 5.9	\\
FreeAnchor\cite{Freeanchor}	  & ResNet-101 	& 0.542		& 0.685		& 55.1	& 6.0	\\
FCOS\cite{FCOS}	        & ResNet-101	& 0.549		& 0.737		& 50.78	& 6.1	\\
LAD\cite{LAD}	          & ResNet-101	& 0.546		& 0.390 	& 51.10	& 4.7	\\
LD\cite{LD}	          & ResNet-101	& 0.590		& 0.331 	& 51.25	& 4.7	\\
PVT\cite{PVT}	& Pyramid-Vision-Transformer-S 	& 0.596	& 0.262	& 52.30	& 4.8	\\
RetinaNet\cite{RetinaNet}	    & ResNet-101	& 0.633		& 0.722		& 55.1	& 6.0	\\
RetinaNet\cite{RetinaNet}	    & ResNeXt-101	& 0.638		& 0.770		& 54.74	& 5.8	\\
CE-RerinaNet\cite{CE-RetinaNet}	& ResNeXt-101	& 0.751		& 0.886		& 57.1	& 5.9	\\
\hline 
ALSS-YOLO	  & ALSSNet	    & 0.679		& 0.571		& 1.40	& 240.4	\\
ALSS-YOLO-s	  & ALSSNet   	& 0.714		& 0.524		& 2.18	& 224.8	\\
ALSS-YOLO-m	  & ALSSNet   	& 0.803		& 0.506		& 2.70	& 219.3\\
\hline 
\end{tabular}\\
\end{table}

\section{Conclusions}

This study introduces ALSS-YOLO, a lightweight detector optimized for UAV TIR images, designed to address challenges in detecting blurred small targets. ALSS-YOLO integrates several key innovations, including the ALSS module, which enhances feature extraction through adaptive channel splitting, bottleneck operations, and depthwise convolution. The model's LCA module encodes global spatial information, and the single-channel focusing module improves feature extraction in TIR images. Additionally, modifications to the localization loss function enhance accuracy for small targets.

Experimental results on the BIRDSAI dataset demonstrate that ALSS-YOLO achieves a 1.7\% higher mAP0.50 compared to the YOLOv8-n' baseline, with the ALSS-YOLO-s and ALSS-YOLO-m models showing further improvements in mAP score and parameter efficiency. These results highlight the model's effectiveness in challenging UAV TIR imaging scenarios, meeting the precision requirements for UAV applications.

Future work will focus on optimizing the ALSS module by exploring dynamic channel division ratios and improving channel shuffling strategies to enhance model performance. Additionally, evaluations will be expanded to include other datasets, and real-world tests will be conducted on UAV platforms to assess the model's applicability in dynamic environments.

\bibliographystyle{IEEEtran}
\bibliography{references}

\begin{IEEEbiographynophoto}{Ang He}
is currently working toward Master's degree in Optoelectronic Information Engineering at the School of Physics and Optoelectronic Engineering, Guangdong University of Technology, and the Guangdong Provincial Key Laboratory of Sensing Physics and Application Integration.  His research interests include remote sensing image processing, artificial intelligence, and deep learning.
\end{IEEEbiographynophoto}

\begin{IEEEbiographynophoto}{Xiaobo Li}
is currently working toward Master's degree in Optoelectronic Information Engineering at the School of Physics and Optoelectronic Engineering, Guangdong University of Technology. His research interests include robot motion control and semi-supervised learning.
\end{IEEEbiographynophoto}

\begin{IEEEbiographynophoto}{Ximei Wu}
is currently working toward Master's degree in Optoelectronic Information Engineering at the School of Physics and Optoelectronic Engineering, Guangdong University of Technology. His research interests include image preprocessing, feature matching, and deep learning.
\end{IEEEbiographynophoto}
  
\begin{IEEEbiographynophoto}{Chengyue Su}
received his Ph.D. degree in Physics and is currently a Professor and Master's Supervisor at the School of Physics and Optoelectronic Engineering, Guangdong University of Technology, where he also serves as the Associate Dean. He is a member of the University Laboratory Construction Committee and the Electrical and Electronic Experimental Teaching Steering Committee. He has participated in two projects funded by the National Natural Science Foundation of China and one provincial science and technology project. His current research interests include robotic motion control and computer vision.
\end{IEEEbiographynophoto}

\begin{IEEEbiographynophoto}{Jing Chen}
is an Associate Professor and Master's Supervisor in the Department of Electronics at Sun Yat-sen University. She received her Ph.D. in Communication and Information Systems from Sun Yat-sen University. She is a recipient of the university-level award under the "Thousand, Hundred, and Ten Talent Project" of Guangdong Province. Her research interests include machine learning, deep learning, 3D object detection, semi-supervised learning, and medical image processing.
\end{IEEEbiographynophoto}

\begin{IEEEbiographynophoto}{Sheng Xu}
was born in Ganzhou, China. He received the M. Eng. degree in Control Engineering and the Ph.D. degree in Control Engineering 
both from the South China University of Techinology Guangzhou China. He once served as an electronic process engineer at BBK Electronics Industry Company and a data analysis engineer at China Unicom. He is currently an assistant professor at Guangdong University of Technology. His research interest include machine vision and motion control.
\end{IEEEbiographynophoto}

\begin{IEEEbiographynophoto}{Xiaobin Guo}
  is a Master's Supervisor who received his Ph.D. degree in June 2018 from the School of Physics and the Key Laboratory for Magnetism and Magnetic Materials of the Ministry of Education, Lanzhou University. He has published over 20 papers and is currently leading one project funded by the National Natural Science Foundation of China for Young Scientists, one Guangzhou Science and Technology Program project, and one university-funded Young Talent Research Initiation project. His primary research interests include magnetic materials and deep learning.
\end{IEEEbiographynophoto}

\end{document}